\documentclass[journal]{IEEEtran}

\usepackage{times}
\usepackage{epsfig}
\usepackage{graphicx}
\usepackage{amsmath}
\usepackage{amssymb}
\usepackage{multirow}
\usepackage{algorithm}
\usepackage{algorithmic}
\usepackage{booktabs}
\usepackage{caption}
\usepackage{makecell}
\usepackage{color}

\usepackage{bbding}
\usepackage{cite}
\usepackage{float} 
\usepackage[numbers]{natbib}

\newlength\myindent
\ifCLASSINFOpdf
\else
\fi
\begin{document}
%
\title{Deep Single Image Deraining using An Asymetric Cycle Generative and Adversarial Framework}
%
%
%

\author{Wei Liu, Rui Jiang, Cheng Chen, Tao Lu and Zixiang Xiong 
\thanks{Wei Liu, Rui Jiang, Cheng Chen and Tao Lu are with the Hubei Key Laboratory of Intelligent Robot, School of Computer Science and Engineering, Wuhan Institute of Technology,
Wuhan 430205, China.
\textit{(Corresponding author: Tao Lu, e-mail: lutxyl@gmail.com)}

Zixiang Xiong is with the Department of Electrical and Computer Engineering, Texas A\&M University, College Station, TX 77843 USA. 
	}
}



%



\maketitle

\begin{abstract}
In reality, rain and fog are often present at the same time, which can greatly reduce the clarity and quality of the scene image.
However, most unsupervised single image deraining methods mainly focus on rain streak removal by disregarding the fog, which leads to low-quality deraining performance. In addition, the samples are rather homogeneous generated by these methods and lack diversity, resulting in poor results in the face of complex rain scenes. To address the above issues, we propose a novel Asymetric Cycle Generative and Adversarial framework (ACGF) for single image deraining that trains on both synthetic and real rainy images while simultaneously capturing both rain streaks and fog features. ACGF consists of a Rain-fog2Clean (R2C) transformation block and a Clean2Rain-fog (C2R) transformation block. 
The former consists of parallel rain removal path and rain-fog feature extraction path by the rain and derain-fog network and the attention rain-fog feature extraction network (ARFE) , while the latter only contains a synthetic rain transformation path.
In rain-fog feature extraction path, to better characterize the rain-fog fusion feature, we employ an ARFE to exploit the self-similarity of global and local rain-fog information by learning the spatial feature correlations. 
Moreover, to improve the translational capacity of C2R and the diversity of models, we design a rain-fog feature decoupling and reorganization network (RFDR) by embedding a rainy image degradation model and a mixed discriminator to preserve richer texture details in synthetic rain conversion path.
Extensive experiments on benchmark rain-fog and rain datasets show that ACGF outperforms state-of-the-art deraining methods. We also conduct defogging performance evaluation experiments to further demonstrate the effectiveness of ACGF.
\end{abstract}

\begin{IEEEkeywords}
Attention mechanism, Unsupervised image deraining, adversarial learning.
\end{IEEEkeywords}

%
\IEEEpeerreviewmaketitle

\section{INTRODUCTION}
Rain is a common weather phenomenon, which significantly affects the visibility and performance of many computer vision tasks, such as object detection \cite{Redmon2016} and segmentation \cite{Li2017}. In a rainy image, there are two main factors affecting the degradation of visibility: rain streaks and fog \cite{Hu2019}, \cite{Li2019a}, \cite{Yang2017}. Thus, based on the foggy image degradation model \cite{Nayar1999} (i.e., atmospheric degradation model), it can be mathematically modeled for an observed rainy image as \cite{Li2019a}, \cite{Yang2016}:
\begin{figure}[ht]
	\begin{center}
		\begin{tabular}{@{}cc@{}}
			\includegraphics[width =0.5 \textwidth]{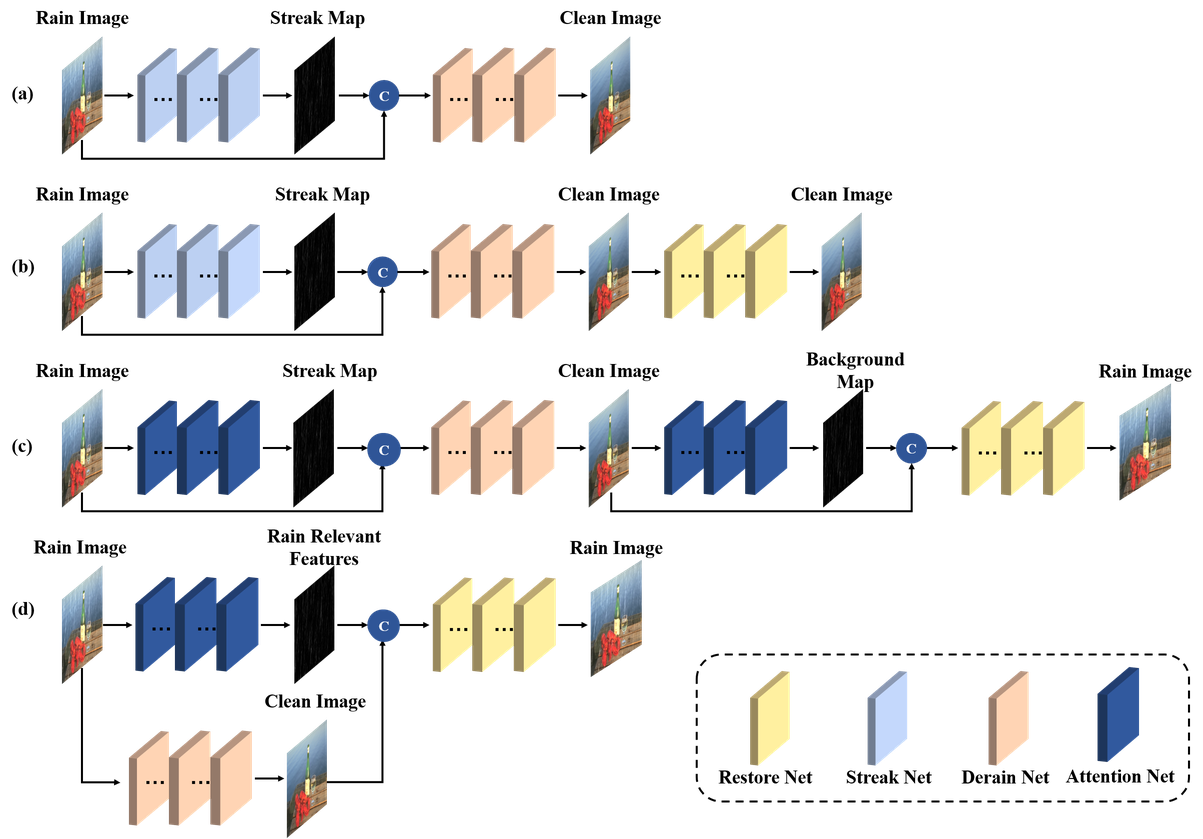}\\
		\end{tabular}
	\end{center}
	\caption{Different learning-based structures for single image deraining. (a) Supervised learning structure. (b) Apply the structure of (a) to unsupervised learning framework \cite{Zhu2017}. (c) Add an extraction of background to the structure (b), e.g. CycleDerain \cite{Wei2021}. (d) Our structure.}
	\vspace{-0.6cm}
	\label{firs-structure} 
\end{figure}

\begin{equation}
	I = t(x)(J(x)+ \sum_{i}^n R_i)  + A(1-t(x)) 
	\label{eq1}
\end{equation}
where $ I $ is the rainy image, $ t(x) $ is the transmission map and $ A $ is the global atmospheric light of the scene. $ J(x) $ is a clear image, $ R_i $ represents the rain streaks at the $ i $-th layer along the line of sight.
Therefore, it is necessary to estimate the atmospheric light $ A $ and transmission $ T $ as well as $ R_i $ to use this model to restore $ J $ from rain-fog images through formulation (1).

Many existing methods \cite{Kang2011, Li2016, Fu2017a, Zhang2018a, Li2018} have bad deraining performance because they ignore the presence of fog. To address this problem, a series of deep-learning-based rain removal methods \cite{Yang2017}, \cite{Li2019a} have proposed. 
They estimate $ A $, $ T $, and $ R_i $ by using supervised learning framework, which train on synthetic paired rain-rainfree images. Although these methods consider both the effect of rain streaks and fog and have a good performance, they may fail to remove the rain from the real-world rainy images. The reason is that rain streak is evenly distributed in the synthetic data, whereas in real-world rain it is not. Recently, some semi-supervised \cite{Wei2020} and unsupervised \cite{Zhu2019}, \cite{Wei2021} learning-based rain removal methods have been proposed to improve the generalization ability of the model in the real world. However, there remain two issues in these methods. First, they still do not take into account the veiling effect caused by fog. 
Second, the existing unsupervised rain removal methods synthesize a homogenous rain image and lack diversity, resulting in these methods having good results for a certain type of rain image but poor results in other rain scenes.

\begin{figure}[tp]
	\begin{center}
		\resizebox{\columnwidth}{!}{
			\begin{tabular}{@{}cccccccccc@{}}
				\includegraphics[width = 0.2\columnwidth]{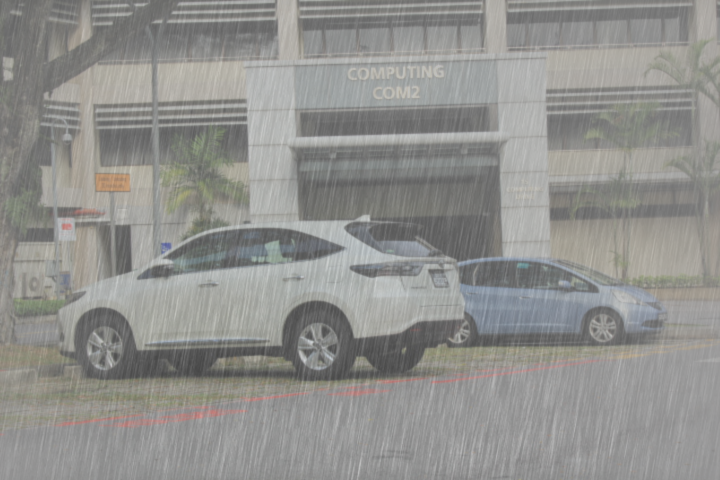}& \hspace{-0.4cm}
				\includegraphics[width = 0.2\columnwidth]{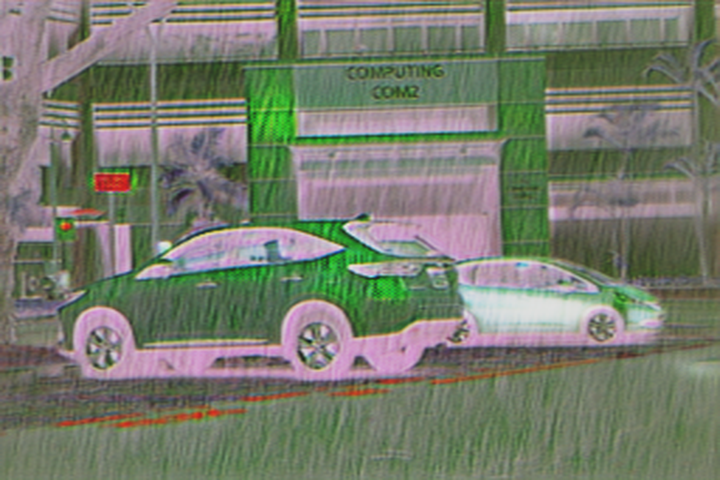} & \hspace{-0.4cm}
				\includegraphics[width = 0.2\columnwidth]{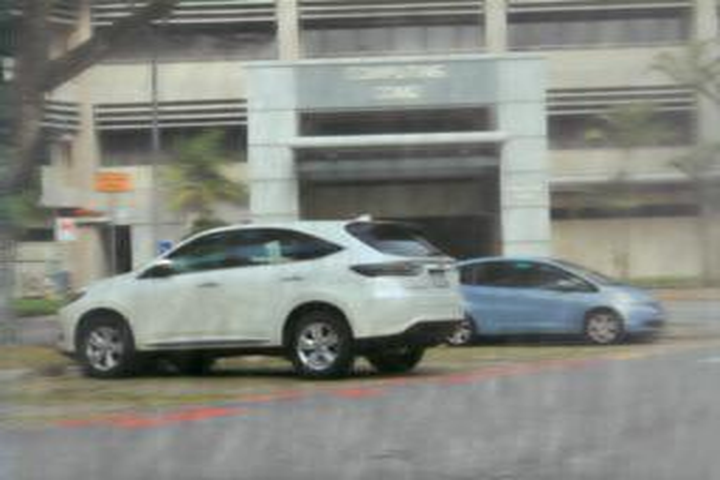}& \hspace{-0.4cm}
				\includegraphics[width = 0.2\columnwidth]{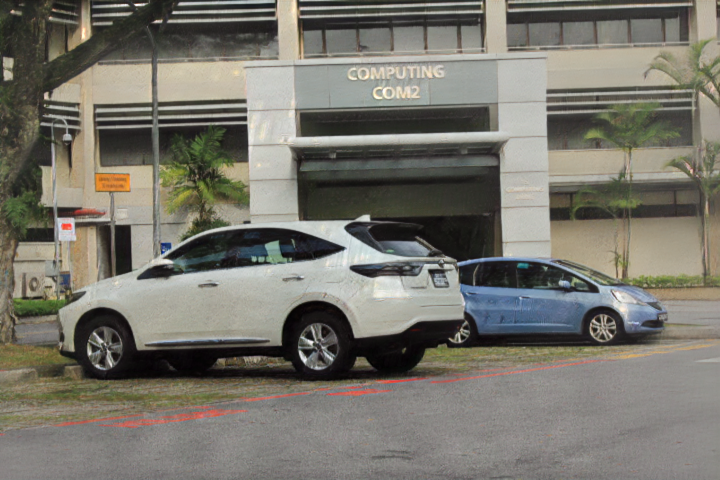}
				\\
				{\tiny (a)} & \hspace{-0.4cm}
				{\tiny (b)} & \hspace{-0.4cm}
				{\tiny (c)} & \hspace{-0.4cm}
				{\tiny (d)} 
				\\
				
			\end{tabular}
		}
	\end{center}
	\vspace{-0.3cm}
	\caption{Images generated by different structures. (a) is the original rain images. (b) is the derained result by the structure of Fig. 1(b). (c) is the derained result by the structure of CycleDerain \cite{Wei2021}. (d) is the derained result by our structure.}
	\vspace{-0.6cm}
	\label{struct compare}
\end{figure}

To address the above limitations, we propose to explore different structures of unsupervised learning-based rain removal methods.
By analyzing and refactoring existing supervised and unsupervised learning derain frameworks, we design a framework that can provide more physical prior knowledge for unsupervised learning and improve model diversity.
At present, as shown in Fig. \ref{firs-structure} (a), most rain removal methods \cite{Ahn2021}, \cite{Li2017a}, \cite{Ye2021} are based on a supervised learning structure. They first design a rain streak extraction network to extract the rain streak feature from the rainy image. They then merge both the streak feature and the rainy image into a derain network to obtain the final clear image. Inspired by this structure, we first apply this manner to an unsupervised learning framework, such as CycleGAN \cite{Zhu2017}. As can be seen from Fig. \ref{firs-structure} (b), to facilitate comparison, we only introduce one transformation path that maps the rainy image domain to the clear image domain. Compared with the supervised learning structure (as shown in Fig. \ref{firs-structure} (a)), this structure only adds a reconstructed network (Restore Network) after the derain network. This structure achieves relatively good results. However, it may cause serious color cast in derained result (as shown in Fig. \ref{struct compare} (b)). To address this problem, a new network structure is proposed \cite{Wei2021} to remove the rain from a single rainy image. Fig. \ref{firs-structure} (c) shows the network structure used by CycleDerain \cite{Wei2021}. As shown in Fig. \ref{firs-structure} (c), the authors of \cite{Wei2021} designed an attention network (which is a Streak net) between the Deraining network and the Restore Network to take care of the rain streak information and background information at the same time. More specifically, this Attention Net is also used to extract a background map, which can guide the Restore Network to reconstruct more details to improve the quality of the derained result generated by the Derain network. This structure can alleviate the color distortion phenomenon. However, it ignores areas visually blocked by the fog and its deraining result has low quality due to blurred edges (as shown in Fig. \ref{struct compare} (c)).
Instead of cascading the Attention and Derain Nets in \cite{Wei2021}, we adopt a parallel structure between them to solve the above problem (as shown in Fig. \ref{firs-structure} (d)). In this structure, the Derain Net is only dedicated to learning the mapping correlations between different types of rainy images and rain-free images. Moreover, our Attention Net is used to extract the rain relevant features, including rain streaks and fog information. Fig. \ref{struct compare} (d) highlights the defogged result generated by our proposed approch, showing better deraining performance than others in visual effect. It is worth noting that in the testing phase, we only need a single network to convert different types of rainy images into clear ones, leading to significant reduction on the number of parameters. 

Based on this asymmetric parallel structure, we propose a novel asymetric cycle generative and adversarial framework (ACGF, as shown in Fig. \ref{totall-structure}), which uses unpaired images in training dataset.
ACGF consists of two transformation blocks: Rain-fog2Clean (R2C) and Clean2Rain-fog (C2R). 
The former learns the mapping relationship between the fog/rain/rain-fog images domain and the clear images domain, and provides prior knowledge for the transformation of the latter from clear images to rain images. The latter uses prior knowledge to convert clear images into fog/rain/rain-fog images, and the two finally complete the mutual conversion from rain images to rain-free images through adversarial learning.
R2C consists of three networks to form two conversion paths: rain-fog removal path and rain-fog feature extraction path.
In it, we propose a novel attention rain-fog feature extraction network (ARFE) by considering both rain streak and fog features and constructing a new deraining network. When extracting rain relevant features from real-world rainy images, ARFE is capable of exploiting the global and local self-similarity of fusion rain-fog information for a deep representation. 
C2R consists of two networks to form a synthetic rain path.
In it, we employ a Rain-fog Feature Decoupling and Reorganization Network (RFDR) to better constrain the deraining network to generate high-quality images. Specially, in FRDR, we utilize a pyramid network by embedding a rain image degradation model to decompose rain-fog features at the same time, leading to the use of high-level semantic information to guide low-level features to perform different feature completions. Furthermore, we propose in ACGF a mixed discriminator to guide the network to distinguish whether a rainy image contains rain or fog relevant features, which can preserve and improve more texture details for deraining result. Specially, we introduce a diverse loss function for the discriminator to learn the difference between images in different image domains to further enhance the diversity of the model. Experimental results on rain-fog dataset, rain dataset, and fog dataset show that ACGF is very competitive in deraining performance and other low-level vision tasks such as single image defogging.  

The rest of the paper is organized as follows. In Section II, we briefly review the existing works of removing rain from a single image. Section III describes the proposed model in detail. Experiments and ablation studies are provided in Section IV. Conclusions are drawn in Section V.

\section{RELATED WORK}
In this section, we briefly review existing models from three categories: model-based methods, learning-based methods, and attention mechanism.
\subsection{Model-based Methods}

Existing model-based methods employ optimization frameworks
for deraining. 
Kang \textit{et al.} \cite{Kang2011} proposed a rain removal framework based on morphological component analysis, which converts the rain removal problem into an image decomposition problem. This method can fully automate and independently decompose rain from images through dictionary learning. Specially, during the dictionary learning stage, no patterns and additional samples are required. 
Chen \textit{et al.} \cite{Chen2014} used guided image filters to decompose the input image into low-frequency and high-frequency parts. By performing dictionary learning and sparse coding, the rainy component of the high-frequency part is removed.
Li \textit{et al.} \cite{Li2016a} applied Gaussian mixture models (GMMs) to model both rain and background layers. The background layer of the GMMs is extracted from real-world scenes with diverse background information.
For the rain layer of the GMMs, they chose a rain patch without background texture from the input image to train.
Zhu \textit{et al.} \cite{Zhu2017a} first analyzed the gradient information in the rain image to determine the image area dominated by rain streaks. Then, they estimate the main rain streak direction from it to extract a series of rain dominated patches. Finally, three special prior knowledge are used to iteratively separate the background details from the rain pattern layer. 
Deng \textit{et al.} \cite{Deng2018} proposed a global sparse model involving three sparse items by considering the inherent direction and structural knowledge of rain patterns and the attributes of image background information.

Although these methods have significant deraining performance, model-based methods still have limitations in preserving rich texture details. For example, they may fail for removing the rain streak from the image captured in a heavy rain condition.

\begin{figure*}[ht]
	\begin{center}
		\vspace{-0.5cm}
		\begin{tabular}{@{}c@{}}
			\includegraphics[width = \textwidth]{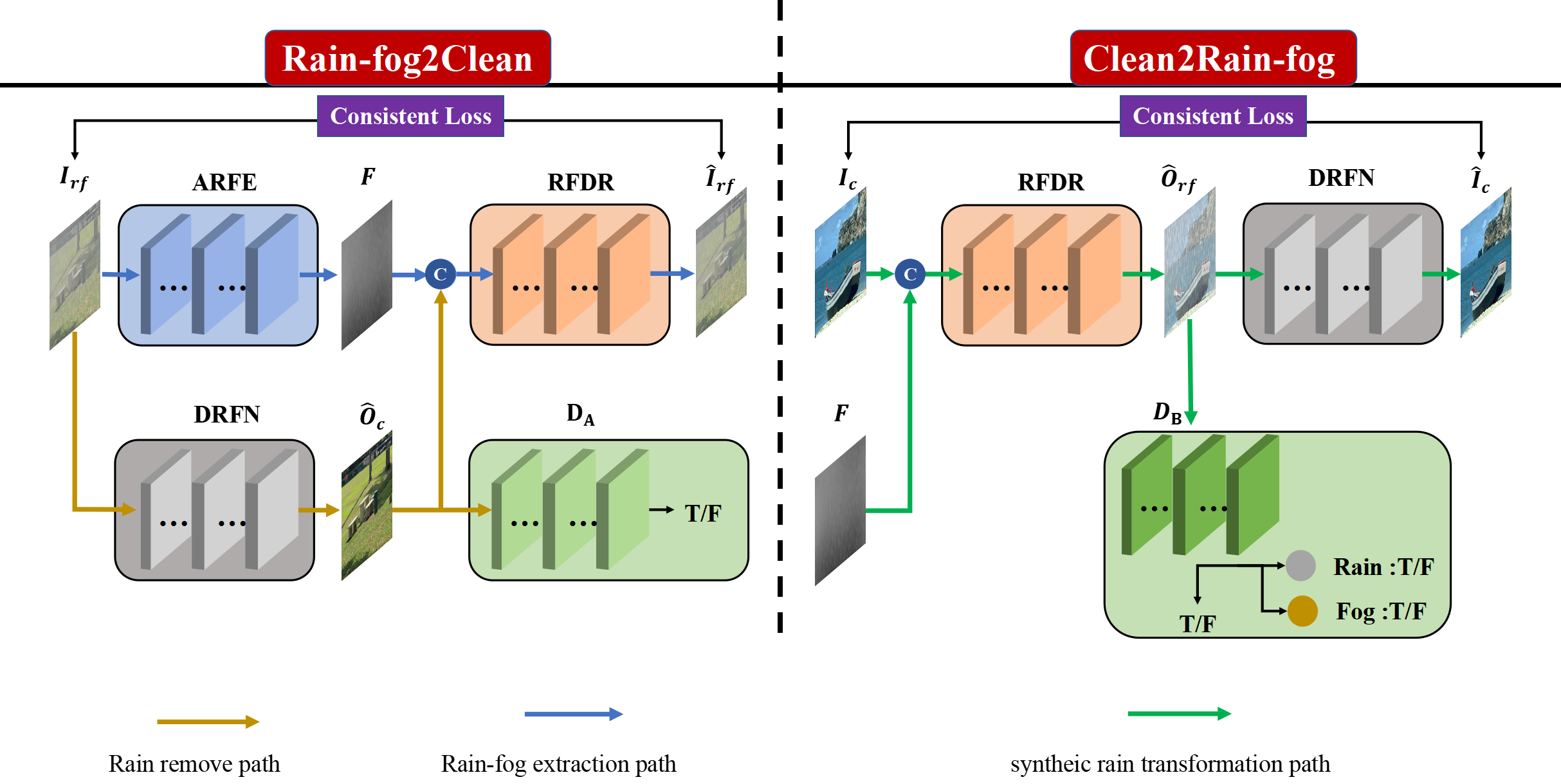}\\
		\end{tabular}
	\end{center}
	\vspace{-0.3cm}
	\caption{The architecture of our proposed ACGF model. 
		The four modules (Derain-fog Network (DRFN); Attention Rain-fog Feature Extraction Network (ARFE); Rain-fog Feature Decoupling and Reorganization Network (RFDR); two discriminators $ D_A $ and $ D_B $.)
		form three transformation paths (Rain remove path; Rain-fog extraction path and syntheic rain transformation path.) and the three transformation paths form two asymmetric blocks Rain-fog2Clean and Clean2Rain-fog.
		In Rain-fog2Clean block,		
		DRFN uses rainy image $ I_{rf} $ as its input and outputs a clear image $ \hat{O}_c $.
		ARFE also takes $ I_{rf} $ as the network’s input to predict the rain-fog related feature $ F $.
		RFDR takes $\hat{O}_{c} $ and $ F $ as input to generate a rainy image $ \hat{I}_{rf} $. $ D_A $ is used to identify the authenticity of the image.
		In Clean2Rain-fog block, the $ F $ extracted in Rain-fog2Clean block and the clear image $ I_c $ are input to RFDR to generate the degraded image $ \hat{O}_{rf} $. Then, DRFN transforms $ \hat{O}_{rf} $ into a clear image $ \hat{I}_c $. Compared with $ D_A $, $ D_B $ adds a branch to determine whether the generated $ \hat{O}_{rf} $ contains rain-fog feature information.
	}
	
	\vspace{-0.5cm}
	\label{totall-structure}
\end{figure*}

\subsection{Learning-based Methods}
\subsubsection{supervised learning for rain streaks removal}
In recent years, many supervised learning-based methods are proposed to remove the rain streaks from a single image. Fu \textit{et al.} designed a convolutional neural network \cite{Fu2017} to learn the mapping function between clean and rainy images. This network only takes high-frequency details as input and predicts the residual rain and clean images. 
To better represent rain streaks, Zhang \textit{et al.} \cite{Zhang2018b} proposed a multi-path densely connected network to automatically detect the rain-density to guide the rain removal.
To reduce the number of network parameters and maintain good deraining performance, Ren \textit{et al.} \cite{Ren2019} proposed a Progressive ResNet (PRN) to remove the rain streaks by using recursive calculations.
To capture visual characteristics beyond signal fidelity, Zhang \textit{et al.} \cite{Zhang2019} use the conditional generative adversarial network (CGAN) for the rain streaks removal. However, during the testing phase, artifacts may be generated when the distribution of the test set and the training set are not the same. Wang \textit{et al.} \cite{Wang2020a} aggregate the advantages of the conventional model-driven prior-based methods and data-driven DL-based methods construct a novel rain convolutional dictionary network (RCDNet) for single image deraining. By using the sparsity and non-local similarity of rain streaks, specific rain cores at different stages are learned to predict rain maps. However, these methods have a low-quality performance when removing rain streaks from a real-world image due to they do not take into account the influence of the fog. 

\subsubsection{supervised learning for rain streaks and fog remove}
To address the problem of the above methods, some supervised learning-based deraining methods by considering the effect of both rain streaks and fog. 
Yang \textit{et al.} \cite{Yang2017}, \cite{Wenhan2020} constructed a new rain model, which adopted multi-stage learning to remove the rain streaks from a single image. More specifically, they used binary graphs to narrow the focus of the network, which achieved a better rain removal effect. However, a large amount of noise hidden in the atmosphere will be enhanced by the binary graphs, which makes the method unable to deal with the fog caused by heavy rain. To address these limitations,
Li \textit{et al.} \cite{Li2019a} proposed a two stage conditional adversarial learning framework to handle the heavy rain image restoration by embedding the physics model. In the first stage, a guided filtering framework is used to decompose the image into high-frequency and low-frequency components to estimate the rain streaks, the transmission map, and the atmospheric light. In the second stage, they designed a depth-guided GAN to restore the background information. 
To handle multiple bad weather degradations with a single network, Li \textit{et al.} \cite{Li2020} designed a generator with multiple task-specific encoders. They first used a neural architecture search to refine the degraded image features underlying different physics principles, which are extracted from all encoders. Then, they designed an adversarial learning scheme based on a multi-task discriminator to classify the degradation type and recover the image under different bad weather conditions. Although these methods pay attention to both the rain streaks and fog, they require a large number of paired rain-rainfree images for training, resulting in long preliminary preparation time and poor generalization. 
\subsubsection{Unsupervised and semi-supervised learning for rain streaks remove}
To improve the generalization of the deraining method on real-world images, Wei \textit{et al.} \cite{Wei2020} proposed a semi-supervised learning-based method. This method provides a dual-path learning paradigm for simultaneously utilizing the supervised and the unsupervised knowledge for the deraining task, which extracts the residual between the input rainy image and its generated clear image to associate with the rain streaks distribution. However, this method did not show an effective rain removal effect, due to the loss function used on the unsupervised path being weak and ill-posed.
To obtain better rain removal performance on real images, Jin \textit{et al.} \cite{Jin2019} proposed an unsupervised deraining generative adversarial network to extract intrinsic priors from unpaired rain and
clean images. Specifically, they designed two collaborative optimized modules, one is used to constrain the difference between the real rain image and the generated rain image, the other is used to ensure the consistency of the background. However, because this method ignores the important information in the clear image and cannot accurately extract the rain streaks, the details will be lost when the dense rain streaks are removed.
To solve this problem, Wei \textit{et al.} \cite{Wei2021} proposed an unsupervised deraining framework termed DerainCycleGAN. By paying attention to the rain image and the clear image information at the same time, the characteristics of the rain image and the clear image are fully utilized to separate the background and rain streaks.
Although these semi-supervised and unsupervised methods have good effects in removing rain streaks, they ignore the fog. The methods they proposed are not very effective in removing the veiling effect caused by heavy rain.

\section{PROPOSED METHOD}
In this section, we present the architecture of our ACGF in detail. 
An overview framework of the proposed ACGF is shown in Fig. \ref{totall-structure}, our ACGF consists of a Rain-fog2Clean block and a Clean2Rain-fog block. In these two blocks, there are five main networks: a Derain-fog network (DRFN, $ G_D $), a Rain-fog Decoupling and Reorganization network (RFDR, $ G_R $), an Attention Rain-fog Feature Extraction network (ARFE, $ G_A $) and two discriminators ($ D_A $ and $ D_B $).

For Rain-fog2clean block, its main purpose is to recover a clear image from the rain image. Given rain image $ I_{rf} $, we first input it into DRFN and ARFE to get the clear image $ \hat{O}_c $ and the rain-fog relevant feature $ F $, respectively. Then we concatenate these two outputs and input them to RFDR to reconstruct the rain image $ \hat{I}_{rf} $. The above process can be expressed by the following formulas:
\begin{equation}
	\begin{cases}
		\hat{O}_c = G_{D}(I_{rf})
		\\
		F = G_{A}(I_{rf})
		\\
		\hat{I}_{rf} = G_{R}(\hat{O}_c, F)
	\end{cases}
\end{equation}

For Clean2Rain-fog block, to improve the RFDR mapping ability to transform the clear image into the rain image, we use the rain-fog relevant feature $ F $ as prior knowledge to guide the network to generate a more natural rain image. As shown in Fig. 3, we combine the clear image $ I_c $ and $ F $ and feed them into RFDR:
\begin{equation}
	\hat{O}_{rf} = G_{R}(I_c, F)
\end{equation}
Then, we put $ \hat{O}_{rf} $ into DRFN to get the clear image $ \hat{I}_c $:
\begin{equation}
	\hat{I}_{c} = G_{D}(\hat{O}_{rf})
\end{equation}
Note that, generate different kinds of rain images is our goal. Therefore, we constrain the RFDR with a mixed discriminator $ D_B $ to generate different kinds of rain images with higher fidelity and texture details.

\subsection{Derain-fog Network}

\begin{figure}[t]
	\begin{center}
		\begin{tabular}{@{}cc@{}}
			\includegraphics[width = 0.5\textwidth]{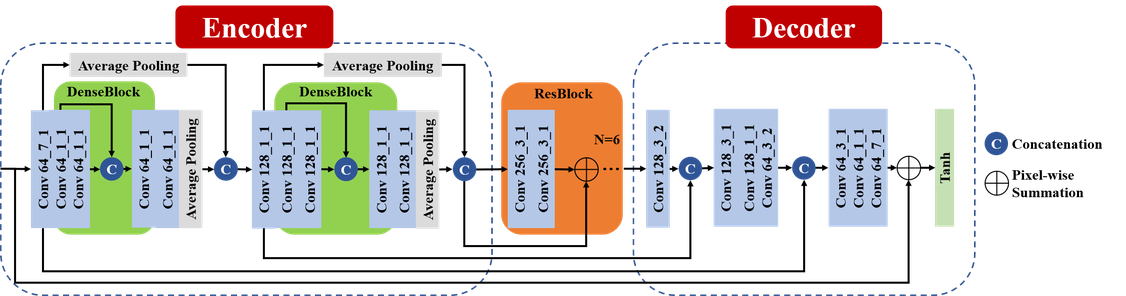}\\
		\end{tabular}
	\end{center}
	\vspace{-0.3cm}
	\caption{ 
		The detail architecture of Derain-fog Network (DRFN). It is an encoder-decoder structure, which uses a densely connection to enable networks to generate the image with high fidelity and rich texture details.
	}
	\vspace{-0.5cm}
	\label{G1-structure}
\end{figure}

\begin{table*}[t]
	\begin{center}
		\caption{
			\\C{\scriptsize ONFIGURATIONS} {\scriptsize OF} {\scriptsize THE} D{\scriptsize ERAIN}-F{\scriptsize OG} N{\scriptsize ETWORK}. {\scriptsize THE} F{\scriptsize EATURE} M{\scriptsize APS} {\scriptsize ARE} D{\scriptsize OWNSAMPLED} {\scriptsize BY} AVG-POOLING {\scriptsize AND} U{\scriptsize PSAMPLED} {\scriptsize BY} T{\scriptsize RANSPOSED} C{\scriptsize ONVOLUTION}.
		}
		\vspace{-0.1cm}
		\resizebox{\textwidth}{!}
		{       \begin{tabular}{|c|c|c|c|c|c|c|c|c|c|c|c|c|c|c|c|c|}
				\hline 
				&
				\multicolumn{8}{c|}{Encoder} 
				&
				\multicolumn{1}{c}{} 
				&
				\multicolumn{7}{|c|}{Decoder} 
				
				\\
				
				\hline
				\textit{layer} & conv1 & denseblock & conv2 & AvgPooling & conv3 & denseblock & conv4 & AvgPooling & residual block & deconv5 & conv6 & conv7 & deconv8 & conv9 & conv10 & conv11  \\
				\hline
				\textit{size} & 7 & 1 & 1 & 3 & 3 & 1 & 1 & 3 & 3 & 3 & 3 & 1 & 3 & 3 & 1 & 7  \\
				\hline
				\textit{channel} & 64 & 64 & 64 & 64 & 128 & 128 & 128 & 128 & 256 & 128 & 128 & 128 & 64 & 64 & 64 & 3 \\
				\hline
				\textit{stride} & 1 & 1 & 1 & 2 & 1 & 1 & 1 & 2 & 1 & 2 & 1 & 1 & 2 & 1 & 1 & 1  \\
				\hline
				\textit{pad} & 0 & 0 & 0 & 1 & 1 & 0 & 0 & 1 & 1 & 1 & 1 & 0 & 1 & 1 & 0 & 0  \\
				\hline
				\textit{sum} &  &  &  &  &  &  & & & & & & & & & & input  \\
				\hline
				\textit{concat} &  &  &  & $ \mathcal{P} $(conv1) &  &  &  & $ \mathcal{P} $(conv3) & & conv3 & & & conv1 & & &   \\
				\hline
			\end{tabular}
		}
		\label{G_layer}   
		\vspace{-0.5cm}
	\end{center} 
\end{table*}

As can be seen from Fig. \ref{G1-structure}, we show the detail architecture of the Derain-fog Network.
In the encoder process, the input image $ I_{rf} $ is first subjected to 7$ \times $7 convolution to extract shallow features. The shallow features can be expressed as:
\begin{equation}
	\mathcal{F}_{s} ^1 =  \varphi_{7 \times 7}(I_{rf})
\end{equation}
where $ \mathcal{F}_{s} ^1 $ denotes the shallow features of the first layer, $ \varphi_{7 \times 7} $ represents a 7$ \times $7 convolution operation. 
Then we use a denseblock to extract contextual semantic information $ \mathcal{F}_{ds} ^1 $ from $ \mathcal{F}^1_s $. Finally, we downsample $ \mathcal{F}^1_s $ and $ \mathcal{F}^1_{ds} $ with an average pooling operation. Thus, we get a first encoding feature $ \mathcal{F}^1_{Enc} $ by concatenating the $ \mathcal{F}^1_s $ and $ \mathcal{F}^1_{ds} $:
\begin{equation}
	\mathcal{F}_{Enc} ^ 1 = [\mathcal{P}(\varphi_{1\times1}(\mathcal{F}_{ds} ^1)), \mathcal{P}(\mathcal{F}_{s} ^1)]
\end{equation}
where $ \mathcal{P} $ represents average pooling operation, $ \varphi_{1 \times 1} $ represents a 1$ \times $1 convolution operation, $ [.] $ is the concatenate operation. Note that, the above procedure is performed twice in Encoder stage, resulting in the output image being 1/4 of the original input:
\begin{equation}
	\begin{cases}
		\mathcal{F}_{s} ^2 = \varphi_{1 \times 1}(\mathcal{F}_{Enc} ^ 1)
		\\
		\mathcal{F}_{Enc} ^ 2 = [\mathcal{P}(\mathcal{F}_{s} ^2), \mathcal{P}(\varphi_{1\times1}(\mathcal{F}_{ds} ^2))]
	\end{cases}
\end{equation}
where $ \mathcal{F}_{s} ^2 $ is the shallow features of the second layer, $ \mathcal{F}_{Enc} ^ 2 $ is second encoding feature.
To avoid the problem of gradient vanishing and overfitting caused by increasing the network depth, we use the residual block to further refine the
encoding feature $ \mathcal{F}_{Enc}^2 $ to $ \mathcal{F}_{res} $. 
In the decoder process, to preserve the image content and rich details from the input, we use a densely connection as follows:
\begin{equation}
	\begin{cases}
		\mathcal{F}_{cat} ^1 = [\gamma_{s2}(\mathcal{F}_{res}), \mathcal{F}_{s} ^2]
		\\
		\mathcal{F}_{cat} ^ 2 = [ \varphi_{1\times1}(\varphi_{1\times1}(\gamma_{s2}((\mathcal{F}_{cat} ^1)))), \mathcal{F}_{s} ^1]
	\end{cases}
\end{equation}
where $ \gamma_{s2} $ represents transposed convolution operation with a step size of 2, $ \mathcal{F}^i_{cat} $ is the $ i $-th densely connection fusion feature. In the last layer, we use the densely connection operation to retain the color information of the image to maintain the structural consistency. The output $ \mathcal{F}_{Dec} $ of the decoder can be expressed as:
\begin{equation}
	\mathcal{F}_{Dec} = \theta (\varphi(\varphi(\varphi_{7 \times 7}(\mathcal{F}_{cat}^2))) + I_{rf})
\end{equation}
where $ \theta $ denotes the Tanh activation function. The specific parameter information of the DRFN is shown in Table \ref{G_layer}.

\subsection{Attention Rain-fog Feature Extraction Network}

\begin{figure}[htbp]
	\begin{center}
		\begin{tabular}{@{}cc@{}}
			\includegraphics[width = 0.5\textwidth]{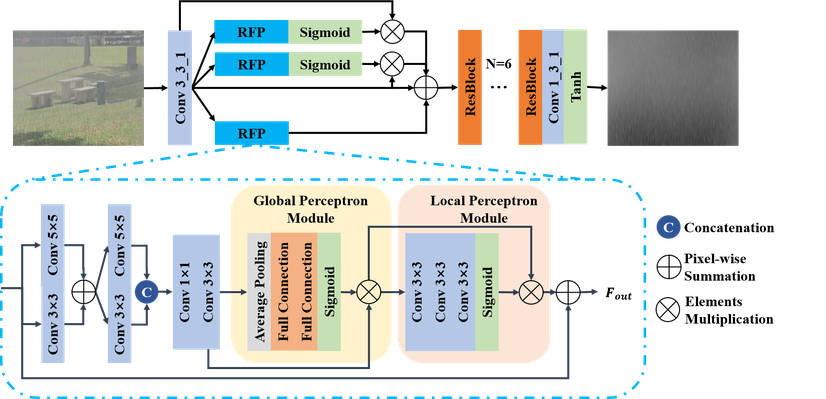}\\
		\end{tabular}
	\end{center}
	\vspace{-0.3cm}
	\caption{ 
		The detail architecture of the Attention Rain-fog Feature Extraction network (ARFE).
		ARFE is mainly composed of rain-fog perception (RFP) blocks and Residual blocks.
		The RFP block use the global and local consistency of rain-fog to model the rain-fog characteristics, which can better extract the rain-fog features (see Fig. \ref{rain-streaks}).
	}
	\vspace{-0.35cm}
	\label{A-structure}
\end{figure}

\begin{table*}[t]
	\begin{center}
		\caption{
			\\C{\scriptsize ONFIGURATIONS} {\scriptsize OF} {\scriptsize THE} ARFE. {\scriptsize IT} C{\scriptsize ONTAINS} N{\scriptsize INE} C{\scriptsize ONVOLUTION}, T{\scriptsize WO} F{\scriptsize ULLY} C{\scriptsize ONNECTED}
			{\scriptsize AND} A{\scriptsize N} A{\scriptsize VERAGE} P{\scriptsize OOLING} O{\scriptsize PERATION}}
		\scalebox{0.8}{
			\begin{tabular}{|c|c|c|c|c|c|c|c|c|c|c|c|c|c|}
				\hline 
				&
				\multicolumn{6}{c|}{} &
				\multicolumn{3}{c|}{Global perceptron} &
				\multicolumn{3}{c|}{Local perceptron} &
				\multicolumn{1}{c|}{}
				
				\\
				
				\hline
				\textit{layer} & conv1 & conv2 & conv3 & conv4 & conv5 & conv6 & AvgPool & linear1 & linear2+S & conv7 & conv8 & conv9+S & output \\
				\hline
				\textit{size} & 3 & 5 & 3 & 5 & 1 & 3 & 1 &  &  & 3 & 3 & 3 & \\
				\hline
				\textit{channel} & 32 & 32 & 32 & 32 & 32 & 32 &  & 8 & 32 & 8 & 8 & 1 & 3 \\
				\hline
				\textit{stride} & 1 & 1 & 1 & 1 & 1 & 1 &  &  &  & 1 & 1 & 1 &\\
				\hline
				\textit{pad} & 1 & 2 & 1 & 2 & 1 & 1 &  &  &  & 1 & 1 & 1 & \\
				\hline
				\textit{sum} &  & conv1 &  &  &  &  &  &  & & & & &input \\
				\hline
				\textit{concat} &  &  &  & conv3 &  &  &  &  &  &  &  & &  \\
				\hline
				\textit{mcl} &  &  &  &  &  &  &  &  & conv6 &  &  & linear2 & \\
				\hline
			\end{tabular}
		}
		\label{ARFE_layer}   
		\vspace{-0.35cm}
	\end{center} 
\end{table*}

To strengthen the attention of both the rain and fog information, we design a ARFE to simultaneously estimate the rain and fog relevant features.
Similar to DRFN, ARFE takes $ I_{rf} $ as the input image, and gets the high-dimensional feature $ \mathcal{F}_e $ by extending the number of channels in the convolutional layer. The $ \mathcal{F}_e $ can be expressed as:
\begin{equation}
	\mathcal{F}_{e} = \varphi_{1 \times 1}(I_{rf})
\end{equation}

\begin{figure}[tp]
	\begin{center}
		\vspace{0.3cm}
		\resizebox{0.75\columnwidth}{!}{
			\begin{tabular}{@{}cccccccccc@{}}
				\includegraphics[width = 0.2\columnwidth]{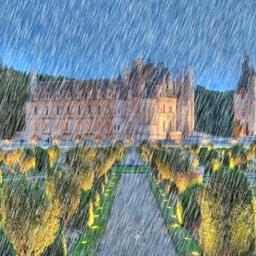}& \hspace{-0.4cm}
				\includegraphics[width = 0.2\columnwidth]{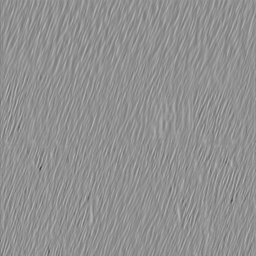}& \hspace{-0.4cm}
				\includegraphics[width = 0.2\columnwidth]{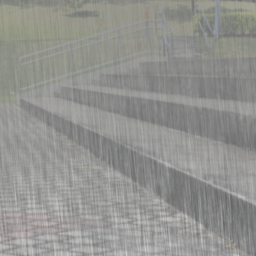}& \hspace{-0.4cm}
				\includegraphics[width = 0.2\columnwidth]{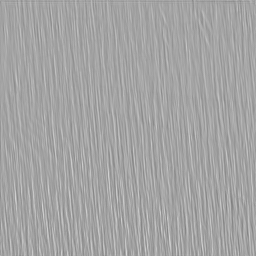}
				\\       
				\includegraphics[width = 0.2\columnwidth]{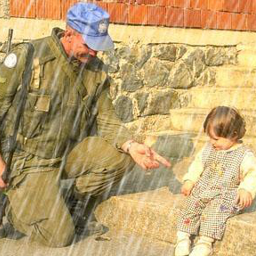}& \hspace{-0.4cm}
				\includegraphics[width = 0.2\columnwidth]{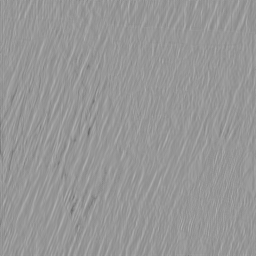}& \hspace{-0.4cm}
				\includegraphics[width = 0.2\columnwidth]{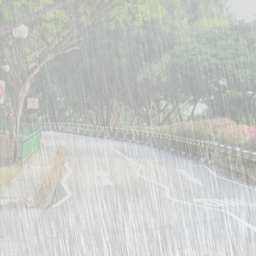}& \hspace{-0.4cm}
				\includegraphics[width = 0.2\columnwidth]{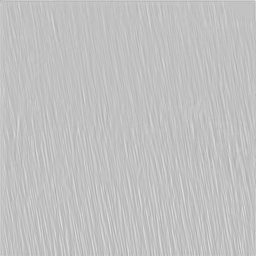}	
				\\
				(a) & \hspace{-0.4cm}
				(b) & \hspace{-0.4cm}
				(c) & \hspace{-0.4cm}
				(d) & \hspace{-0.4cm}
				
				\\
				
			\end{tabular}
		}
	\end{center}
	\vspace{-0.4cm}
	\caption{Feature maps extracted from rain image and rain-fog images. (a) Rain image with different types of rain streaks. (b) The extracted rain streak maps from (a). (c) Rain-fog image with different types of rain streaks. (d) The extracted fusion rain-fog maps from (c).}
	\vspace{-0.5cm}
	\label{rain-streaks}
\end{figure}

Then we use several multiple parallel RFP blocks to aggregate global rain and fog information and be specialized in different feature scales.
Specifically, we introduce a global perception module (GPM) and a local perception module (LPM) to enhance the discriminative learning ability of the network to express rain and fog feature more accurately by focusing on the most informative knowledge.
In the first layer of RFP block (as shown in Fig. \ref{A-structure}), we use 3$ \times $3 and 5$ \times $5 convolution kernels to learn the multi-scale correlations of the rain and fog relevant features. This process can be expressed as:
\begin{equation}
	\begin{cases}
		\mathcal{F}_{ms}^1 = \varphi_{3 \times 3}(\mathcal{F}_e) + \varphi_{5 \times 5}(\mathcal{F}_e)
		\\
		\mathcal{F}_{ms}^2 = [\varphi_{3 \times 3}(\mathcal{F}_{ms}^1) , \varphi_{5 \times 5}(\mathcal{F}_{ms}^1)]
		\\
		\mathcal{F}_{mul} = \varphi_{3 \times 3}(\varphi_{1 \times 1}(I_{rf}))  
	\end{cases}
\end{equation}
where, $ \mathcal{F}_{ms}^i $ is the $ i $-th multi-scale feature fusion information, $ \mathcal{F}_{mul} $ is the final multi-scale fusion feature.
Thereafter, we use GPM to assign weights to
different channels of $ \mathcal{F}_{mul} $, and further model the long-distance dependence and position mode of rain streaks and fog features. The extracted global features $ \mathcal{F}_{global}^i $ can be expressed as:
\begin{equation}
	\mathcal{F}_{global}^i =  R(\sigma(\delta (\delta (\mathcal{P}(\mathcal{F}_{mul})))) * \mathcal{F}_{mul}
\end{equation}
where $ \delta $ denotes the fully-connected layers, $ R $ means the resize operation. $ \sigma $ refers to the sigmoid operation. To refine the correlated information for a better fusion and presentation, we exploit LPM to perform weight assignments on global features. The extracted local features $ \mathcal{F}_{local}^i $ can be expressed as:
\begin{equation}
	\mathcal{F}_{local}^i = \sigma(\varphi_{3\times3}(\varphi_{3\times3}(\varphi_{3\times3}(\mathcal{F}_{global}^i)))) * \mathcal{F}_{global}^i
\end{equation}
Combining the global representational capacity and positional perception of GPM with the local prior of LPM can improve the performance of the extracted rain-fog feature.
The output of the RFP block is formulated as:
\begin{equation}
	\mathcal{F}_{out}^i = \mathcal{F}_e + \mathcal{F}_{local}^i 
\end{equation}
To effectively highlight the rain and fog related features in the spatial dimension. We use a sigmoid function to weight the output of the block to adjust the importance of the information at different spatial positions of the original input $ \mathcal{F}_e $. Then we get the final fusion feature $ \mathcal{F}_{fus} $ through a certain fusion strategy:
\begin{equation}
	\mathcal{F}_{fus} = \sum_{i=1}^{2}\sigma (\mathcal{F}_{out}^i) + \mathcal{F}_{out} ^ 3 + \mathcal{F}_{e}
\end{equation}
where $ \mathcal{F}_{out}^i $ is the output of the $ i $-th RFP block. Finally, several residual blocks are used to further refine this fusion feature to extract rain-fog relevant features. As shown in Fig. \ref{rain-streaks}, our method can effectively extract the rain streak feature, as well as capture the fusion rain-fog feature. It can be clearly seen that the contrast of the latter is significantly higher than the former due to the influence of the fog, which indicates our method is sensitive to the fog. The specific parameters are shown in Table \ref{ARFE_layer}.

\subsection{Rain-fog Feature Decoupling and Reorganization Network}
\begin{figure}[htbp]
	\begin{center}
		\begin{tabular}{@{}cc@{}}
			\includegraphics[width = 0.5\textwidth]{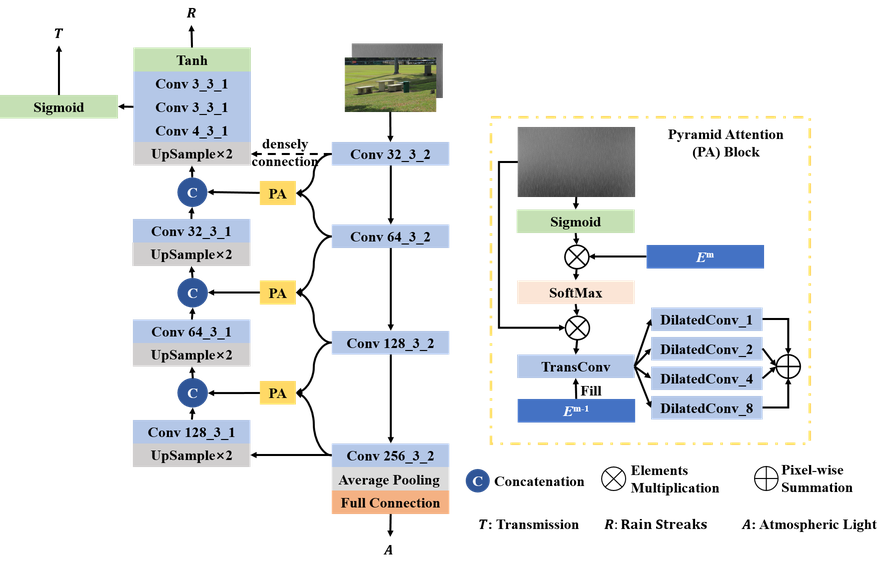}\\
		\end{tabular}
	\end{center}
	\vspace{-0.3cm}
	\caption{The detail architecture of Rain-fog Feature Decoupling and Reorganization network (RFDR), which aggregates the most useful information from the high-level pyramid to the bottom-level pyramid by the Pyramid Attention (PA) block to reduce the influence of background information and feature redundancy.
	}
	\vspace{-0.5cm}
	\label{F-structure}
\end{figure}

\begin{table*}[t]
	\begin{center}
		\caption{
			\\C{\scriptsize ONFIGURATIONS} {\scriptsize OF} {\scriptsize THE} RFDR N{\scriptsize ETWORK}. U{\scriptsize PWARD} A{\scriptsize RROW} I{\scriptsize NDICATES} {\scriptsize THE} U{\scriptsize PSAMPLING} O{\scriptsize PERATION}}
		\scalebox{0.8}{
			\begin{tabular}{|c|c|c|c|c|c|c|c|c|c|c|}
				\hline 
				\multicolumn{1}{|c}{} &
				\multicolumn{2}{|c|}{} &
				\multicolumn{2}{c|}{PA block3} &
				\multicolumn{6}{c|}{}
				\\				\cline{2-5}
				
				\multicolumn{1}{|c}{} &
				\multicolumn{1}{|c|}{} &
				\multicolumn{2}{c|}{PA block2} &
				\multicolumn{1}{c|}{} &
				\multicolumn{6}{c|}{} 
				\\				\cline{2-5} 
				&
				\multicolumn{2}{c|}{PA block1} &
				\multicolumn{2}{c|}{} &
				\multicolumn{6}{c|}{} 
				\\
				\hline
				
				\textit{layer} & conv1 & conv2 & conv3 & conv4 & $ \uparrow $ + conv5 &  $ \uparrow $ + conv6 & $ \uparrow $ + conv7 & $ \uparrow $ + conv8 & conv9 & conv10 \\
				\hline
				\textit{size} & 3 & 3 & 3 & 3 & 3 & 3 & 3 & 3 & 3 & 3     \\
				\hline
				\textit{channel} & 32 & 64 & 128 & 256 & 128 & 64 & 32 & 32 & 4 & 3 \\
				\hline
				\textit{stride} & 2 & 2 & 2 & 2 & 1 & 1 & 1 & 1 & 1 & 1  \\
				\hline
				\textit{pad} & 1 & 1 & 1 & 1 & 1 & 1 & 1 & 1 & 1 & 1 \\
				\hline
				\textit{concat} &  &  &  &  & PA block3 & PA block2 & PA block1 & & & \\
				\hline
				\textit{sum} &  &  &  &  &  &  & input &  & & \\
				
				\hline
				
		\end{tabular} }
		\label{F_layer}   
		\vspace{-0.5cm}
	\end{center} 
\end{table*}

To reduce the influence of the background information contained in the rain and fog features on the subsequent rain-fog image generation.
We designed a feature pyramid inspired by \cite{Zeng2019}. It decomposes the rain-fog features at the same time by using the high-level semantic information to guide low-level features to perform different feature completions. As shown in Fig. \ref{F-structure}, for input rain-fog related feature ($ F $) and clear image ($ I_c $), the output of the $ m $-th layer in pyramid-context encoding stage is $ E^m $. The encoding stage is defined as:
\begin{equation}
	\begin{cases}
		E^1 = \varphi_{3 \times 3}([F, I_c])\\
		\qquad \vdots \\
		E^m = \varphi_{3 \times 3}(E^{m-1})
	\end{cases}
\end{equation}
In this work, $ m=4 $.
Then, we exploit three Pyramid Attention (PA) blocks to calculate the affinity weights between the pyramid features and the rain-fog related feature to remove extra background information to further preserve the rain streaks feature. Specifically, the affinity weights can be expressed as:
\begin{equation}
	w_a = \frac{e^{E^m \odot \sigma(F)}}{\sum_{j=1}^{K} e^{E^m \odot  \sigma(F)}}, j=1,2,\dots,K
\end{equation}
where $ \odot $ refers to the element-wise multiplication operation. $ K $ denotes the total number of channels.
The output of the $ i$-th PA block can be defined as:
\begin{equation}
	\mathcal{F}_{pa}^i = \sum_{n  \in  \left \{1, 2, 4, 8  \right \}}\psi _{n}(\gamma({F}*w_a))
\end{equation}
where, $ \gamma $ is transposed convolution operation. The low-level pyramid features ($ E^{m-1} $) are used as the parameters of this convolution. $ \psi_n $ refer to dilated convolution operation with dilation rate $ n $.
The specific detail parameters are shown in Table \ref{F_layer}.

In practice, however, due to the diversity and complexity of rain-fog distribution, it is very difficult to directly use CNN to synthesis the rain-fog image. Therefore, we introduced the formula (\ref{eq1}) to embed it in our RFDF network. We encode the atmospheric light A at the bottom of the pyramid:
\begin{equation}
	A = \delta(\mathcal{P}(E^4))
	\label{A}
\end{equation}
In pyramid-context decoding stage, the features learned by the PA block will be combined with the bottom-level pyramid features to be upsampled. The decoding stage is divided into four layers of output, and the output of the first layer is:
\begin{equation}
	D^1 = [\varphi_{3 \times 3}(\uparrow (E^4)), \mathcal{F}_{pa}^3]
\end{equation}
where $ \uparrow $ is the upsampling operation.
The output of the second to third layers is:
\begin{equation}
	D^n = [\varphi_{3 \times 3}(\uparrow (D^{n-1})), \mathcal{F}_{pa}^{4-n}], n=2, 3
\end{equation}
where $ D^n $ represents the output of $ n $-th layer in decoding stage. The output of the fourth layer is:
\begin{equation}
	D^4 = \uparrow(D^3)
\end{equation}
Finally, we decode the transmittion ($ T $) and rain streaks ($ R $):
\begin{equation}
	\begin{cases}
		T =\sigma(\varphi_{3 \times 3}(\varphi_{3 \times 3}(D^4)) )
		\\
		R = \theta(\varphi_{3 \times 3}((\varphi_{3 \times 3}(\varphi_{3 \times 3}(D^4)))))
	\end{cases}
	\label{T}
\end{equation}
We input the variables $ A $, $ T $, $ R $ into Eq. (1) to get the final generated rain image ($ \hat{O}_{rf} $).
Note that, in last two deconvolutional block, there
is a densely connection to be added by an element-wise summation from the convolutional, which enforces the network to learn more details. The specific detail parameters are shown in Table. \ref{F_layer}.

\subsection{Discriminator}

\begin{figure}[htbp]
	\begin{center}
		\begin{tabular}{@{}cc@{}}
			\includegraphics[width = 0.5\textwidth]{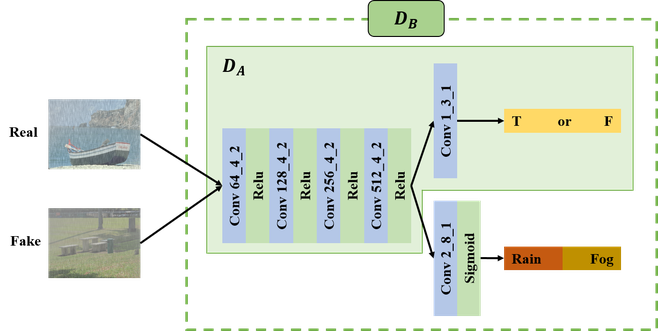}\\
		\end{tabular}
	\end{center}
	\vspace{-0.3cm}
	\caption{
		Structure of the Discriminator. For $ D_B $, we add a new branch on top of $ D_A $ to determine whether the image contains rain or fog features.}
	\vspace{-0.05cm}
	\label{Discriminator-structure}
\end{figure}

We designed two discriminator networks and named them as $ D_A $ and $ D_B $.
$ D_A $ and $ D_B $ have a similar structure as shown in Fig. \ref{Discriminator-structure}. For $ D_A $, we use four-layer convolution and relu activation function to turn the input image into a feature map with the size of the original image 1/16. Each layer has the same kernel size 4 × 4 with a stride of 2, and the filters are 64, 128, 256, 512 from lowest to highest. Finally, we use one convolution with a stride of 1, and the kernel size is 3 × 3 to learn to distinguish between real and fake images.
For $ D_B $, we added a new branch based on $ D_A $. This new branch contains a convolutional layer and a sigmoid activation function, which is used to distinguish whether the image contains the feature of rain or fog. The last convolutional layer is with a kernel size of 8 × 8 and a stride of 1.

\begin{algorithm}[t]
	\caption{Training Process}
	\begin{algorithmic}[1]
		
		\REQUIRE ~~\\ 
		The rain-fog image set $I_{rf}$ ;\\
		The clear image set $I_c$ ;\\
		\ENSURE ~~\\
		The clear image $\hat{O}_{c}$ ;\\
		The rain-fog image $\hat{O}_{rf}$ ; \\
		The rain-fog relevant feature $ F $ ; \\
		\STATE \textbf{Rain-fog2Clean:}
		\begin{ALC@g}
			\FOR{ $x_{rf}$ $\in$ $I_{rf}$, $y_c$ $\in$ $I_c$} 	
			\STATE $F$ = $G_A(x_{rf})$; \\
			\STATE $\hat{O}_{c}$ = $G_D(x_{rf})$; \\
			\STATE $ A_{\hat{O}_{c}} $, $ T_{\hat{O}_{c}} $, $ R_{\hat{O}_{c}} $  $\gets$ $G_R(\hat{O}_{c}, F)$ 
			\\ $\hat{I}_{rf} = T_{\hat{O}_{c}}(\hat{O}_{c} + R_{\hat{O}_{c}}) + A_{\hat{O}_{c}}(1-T_{\hat{O}_{c}})  $
			;\\
			\STATE Adversarial Loss: $min(\parallel D_A(\hat{O}_{c})-1\parallel _2)$;\\
			\STATE Cycle Consistency Loss: $min(\parallel x_{rf}-\hat{I}_{rf}\parallel _1)$;\\
			\STATE Perceptual Loss: $min(\parallel \phi_n(x_{rf}) -\phi_n(\hat{I}_{rf})\parallel _2)$;\\
			\ENDFOR
		\end{ALC@g}	
		\RETURN $\hat{O}_{c}$, $ F $
		\STATE \textbf{Clean2Rain-fog:}
		\begin{ALC@g}
			\FOR{ $y_c$ $\in$ $I_c$, $x_{rf}$ $\in$ $I_{rf}$  } 	
			\STATE  $ A_{y_c} $, $ T_{y_c} $, $ R_{y_c} $ $\gets$ $G_R(y_{c}, F)$\\
			$\hat{O}_{rf}$ = $ T_{y_c}(y_c + R_{y_c}) + A_{y_c}(1-T_{y_c}) $ ; \\
			\STATE  $\hat{I}_{c}$ = $G_D(\hat{O}_{rf})$; \\
			\STATE Adversarial Loss: $min(\parallel D_B(\hat{O}_{rf})-1\parallel _2)$;\\
			\STATE Cycle Consistency Loss: $min(\parallel y_{c}-\hat{I}_{c}\parallel _1)$;\\
			\STATE Perceptual Loss: $min(\parallel \phi_n(y_{c}) -\phi_n(\hat{I}_{c})\parallel _2)$;\\
			\STATE Diverse Loss: $min(\mathbb{E}[logD_B(C=c_i|\hat{O}_{rf})]$;\\
			\ENDFOR
		\end{ALC@g}
		\RETURN $\hat{O}_{rf}$	
	\end{algorithmic}
	\label{restore-algorithm}
\end{algorithm}

\subsection{Loss Function}
The structures of ACGF for single image deraining are trained by four loss functions as follows:
\subsubsection{Adversarial Loss Function}The discriminator is trained to maximize the log-likelihood of the correct source:
\begin{equation}
	\mathcal{L}_{rf}(D_A, G_R, G_A)=\mathop{\mathbb{E}}\limits_{x_{rf}\sim I_{rf}}[log(D_A(x_{rf}))]+
	\mathop{\mathbb{E}}\limits_{y_{c}\sim I_{c}}[log(1-D_A(G_R(G_A(x_{rf}), y_c)))]
\end{equation}
\begin{equation}
	\mathcal{L}_{c}(D_B, G_D)=\mathop{\mathbb{E}}\limits_{y_{c}\sim I_{c}}[log(D_B(y_{c}))] + \mathop{\mathbb{E}}\limits_{x_{rf}\sim I_{rf}}[log(1-D_B(G_D(x_{rf})))]
\end{equation}
where $G_R $, $ G_A$ and $ G_D $ represent RFDR, ARFE and DRFN respectively; $D_A $, $ D_B$ respectively represent two different discriminators, $ I_{rf} $ refer to real rain-fog image, $ I_{c} $ is real clear image. $ \mathcal{L}_{rf} $ fooled the discriminator $ D_A $ by encouraging the DRFN to recover high-quality clean image. 
$ \mathcal{L}_{c} $ aims to fool the discriminator $ D_A $ by making RFDR generate more realistic degraded images containing rain or fog.
To summarize, the adversarial loss function $ \mathcal{L}_{adv} $ for the discriminator is:
\begin{equation}
	\mathcal{L}_{adv} = \mathcal{L}_{rf}(D_A, G_R, G_A)+\mathcal{L}_{c}(D_B, G_D)
\end{equation}

\subsubsection{Cycle Consistency Loss Function}To retain the contents of the generated result consistent with the original image in the same domain, we use a cycle consistency loss function:
\begin{equation}
	\mathcal{L}_{cons}={\left \| x_{rf} - \hat{I}_{rf} \right \| }_2+{\left \| y_{c} - \hat{I}_{c} \right \| }_2
\end{equation}

\subsubsection{Perceptual Loss}
To learn more texture details from the rain image, we introduce the trained VGG19 model \cite{Bermeitinger2017} to the original image and the reconstructed image. This objective is defined as:
\begin{equation}
	\mathcal{L}_{per}={\left \| \phi_n (x_{rf}) - \phi_n(\hat{I}_{rf}) \right \| }_2 + {\left \| \phi_n(y_{c}) - \phi_n(\hat{I}_{c}) \right \| }_2
\end{equation}	
In the formula, $ \phi_n $ represents the characteristics of the $ n $-th layer of the VGG19 network.

\subsubsection{Diverse Loss}
To learn the difference among images in different domains by the discriminator, we introduce a diverse loss function:
\begin{equation}
	\mathcal{L}_{div}=\mathbb{E}[logD_B(C=c_i|\hat{O}_{rf})] + \mathbb{E}[logD_B(C=c_i|I_{rf})]
\end{equation}
where $ \hat{O}_{rf} $ represent the fake rain-fog image and $ I_{rf} $ is real rain-fog image.
Finally, our total loss function for single image deraining with an unsupervised learning-based framework by considering both the rain streaks and fog is expressed as:	
\begin{equation}
	\mathcal{L}_{total}=\lambda_1 \mathcal{L}_{per}+\lambda_2 \mathcal{L}_{cons}+
	\lambda_3\mathcal{L}_{adv} + \lambda_4\mathcal{L}_{div}
	\label{loss}
\end{equation}		
In the formula, $ \lambda_1 $, $ \lambda_2 $, $ \lambda_3 $ and $ \lambda_4 $ are positive weights. The overview of our training procedure for our network is shown in Algorithm \ref{restore-algorithm}.

\section{EXPERIMENTAL RESULTS}
In this section, we qualitatively and quantitatively evaluate our proposed method and compare it with the other state-of-the-arts on synthetic rain-fog datasets, synthetic rain datasets, real-world rain datasets and natural foggy images. 
\begin{figure*}[t]
	\begin{center}
		\resizebox{\textwidth}{!}{
			\begin{tabular}{@{}cccccccccc@{}}
				
				\includegraphics[width = 0.15\textwidth ]{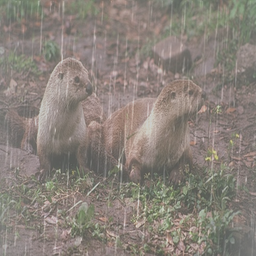}& \hspace{-0.4cm}
				\includegraphics[width = 0.15\textwidth ]{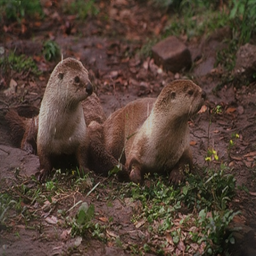}& \hspace{-0.4cm}
				\includegraphics[width = 0.15\textwidth ]{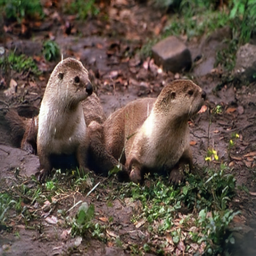}& \hspace{-0.4cm}
				\includegraphics[width = 0.15\textwidth ]{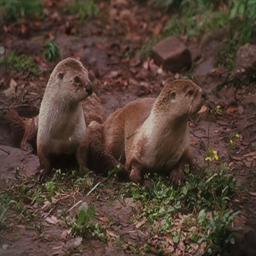}& \hspace{-0.4cm}
				\includegraphics[width = 0.15\textwidth ]{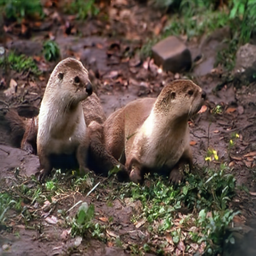}& \hspace{-0.4cm}
				\includegraphics[width = 0.15\textwidth ]{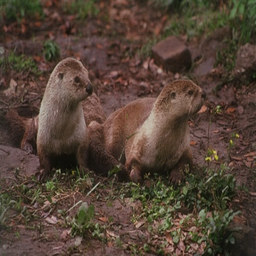}& \hspace{-0.4cm}
				\includegraphics[width = 0.15\textwidth ]{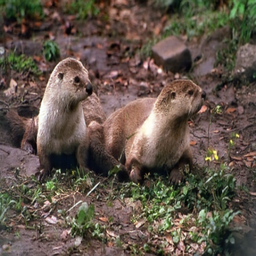}& \hspace{-0.4cm}
				\includegraphics[width = 0.15\textwidth ]{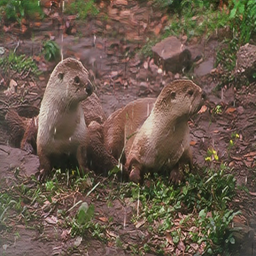}& \hspace{-0.4cm}
				\includegraphics[width = 0.15\textwidth ]{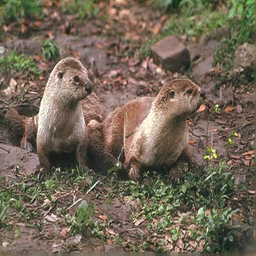}& \hspace{-0.4cm}
				\includegraphics[width = 0.15\textwidth ]{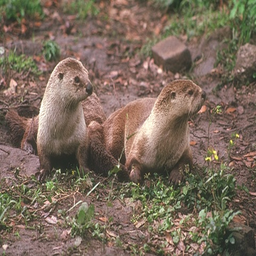}
				
				\\          
				\includegraphics[width = 0.15\textwidth ]{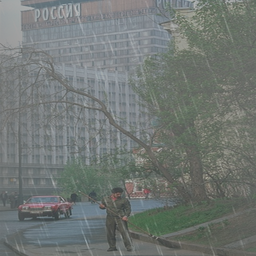}& \hspace{-0.4cm}
				\includegraphics[width = 0.15\textwidth ]{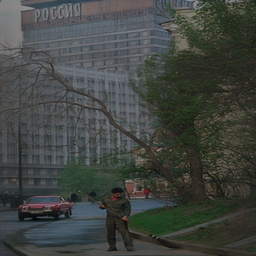}& \hspace{-0.4cm}
				\includegraphics[width = 0.15\textwidth ]{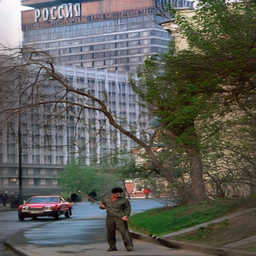}& \hspace{-0.4cm}
				\includegraphics[width = 0.15\textwidth ]{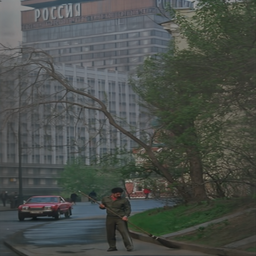}& \hspace{-0.4cm}
				\includegraphics[width = 0.15\textwidth ]{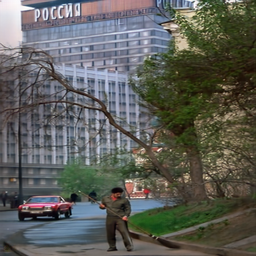}& \hspace{-0.4cm}
				\includegraphics[width = 0.15\textwidth ]{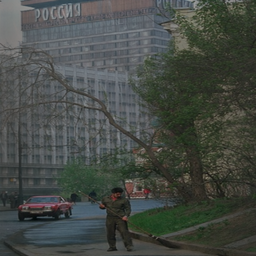}& \hspace{-0.4cm}
				\includegraphics[width = 0.15\textwidth ]{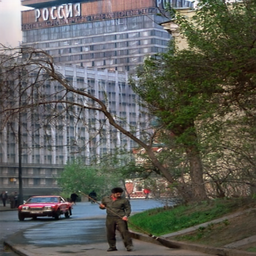}& \hspace{-0.4cm}
				\includegraphics[width = 0.15\textwidth ]{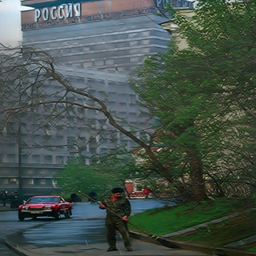}& \hspace{-0.4cm}
				\includegraphics[width = 0.15\textwidth ]{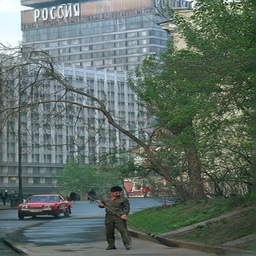}& \hspace{-0.4cm}
				\includegraphics[width = 0.15\textwidth ]{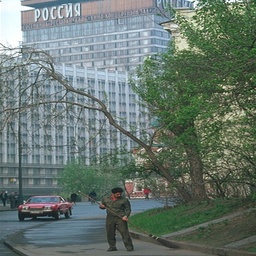}
				
				\\
				\includegraphics[width = 0.15\textwidth ]{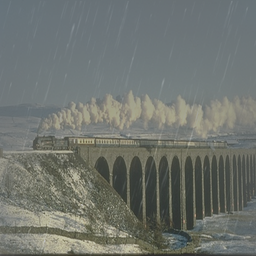}& \hspace{-0.4cm}
				\includegraphics[width = 0.15\textwidth ]{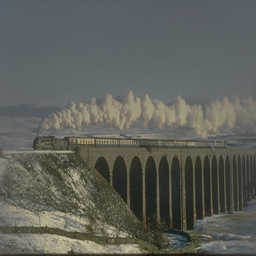}& \hspace{-0.4cm}
				\includegraphics[width = 0.15\textwidth ]{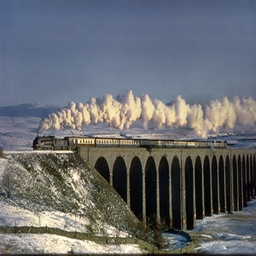}& \hspace{-0.4cm}
				\includegraphics[width = 0.15\textwidth ]{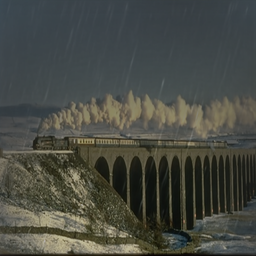}& \hspace{-0.4cm}
				\includegraphics[width = 0.15\textwidth ]{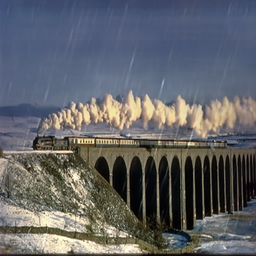}& \hspace{-0.4cm}
				\includegraphics[width = 0.15\textwidth ]{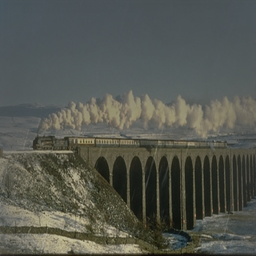}& \hspace{-0.4cm}
				\includegraphics[width = 0.15\textwidth ]{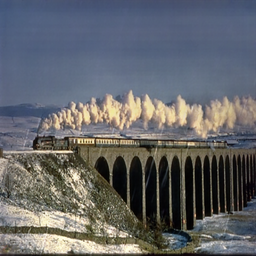}& \hspace{-0.4cm}
				\includegraphics[width = 0.15\textwidth ]{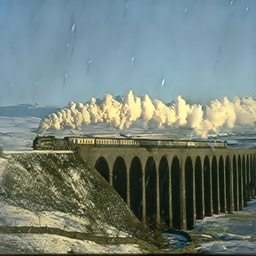}& \hspace{-0.4cm}
				\includegraphics[width = 0.15\textwidth ]{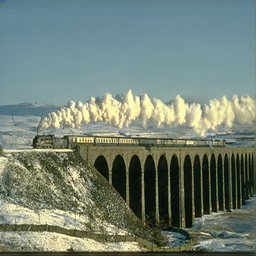}& \hspace{-0.4cm}
				\includegraphics[width = 0.15\textwidth ]{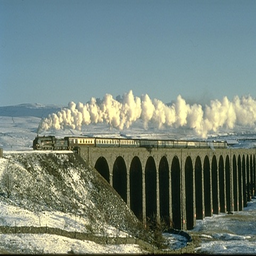}
				\\
				\includegraphics[width = 0.15\textwidth ]{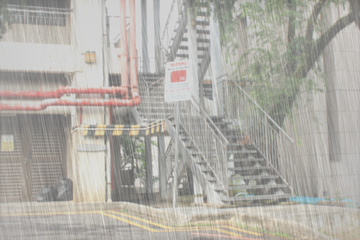}& \hspace{-0.4cm}
				\includegraphics[width = 0.15\textwidth ]{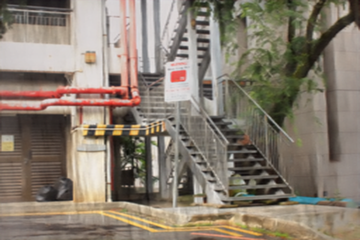}& \hspace{-0.4cm}
				\includegraphics[width = 0.15\textwidth ]{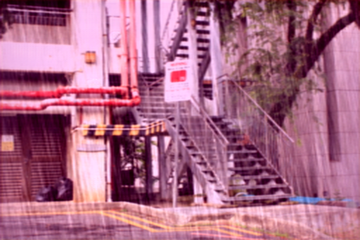}& \hspace{-0.4cm}
				\includegraphics[width = 0.15\textwidth ]{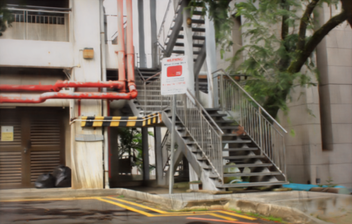}& \hspace{-0.4cm}
				\includegraphics[width = 0.15\textwidth ]{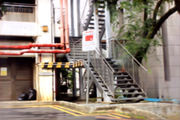}& \hspace{-0.4cm}
				\includegraphics[width = 0.15\textwidth ]{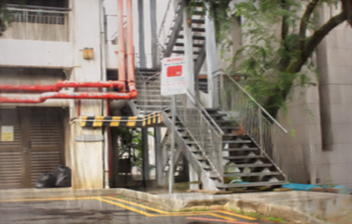}& \hspace{-0.4cm}
				\includegraphics[width = 0.15\textwidth ]{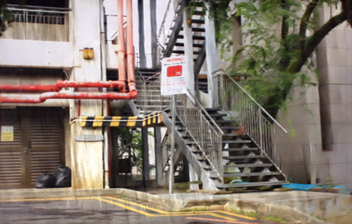}& \hspace{-0.4cm}
				\includegraphics[width = 0.15\textwidth ]{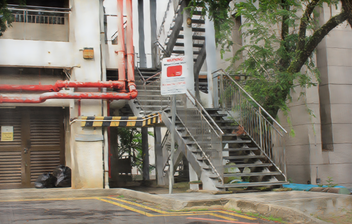}& \hspace{-0.4cm}
				\includegraphics[width = 0.15\textwidth ]{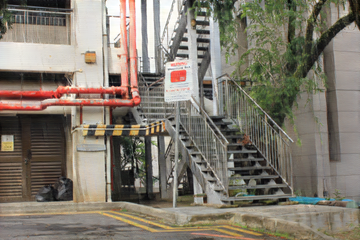}& \hspace{-0.4cm}
				\includegraphics[width = 0.15\textwidth ]{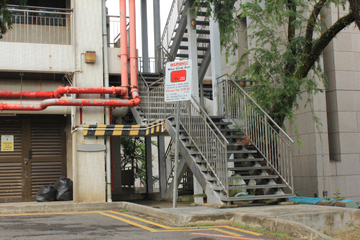}
				\\       
				
				\includegraphics[width = 0.15\textwidth ]{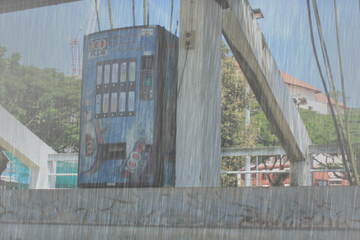}& \hspace{-0.4cm}
				\includegraphics[width = 0.15\textwidth ]{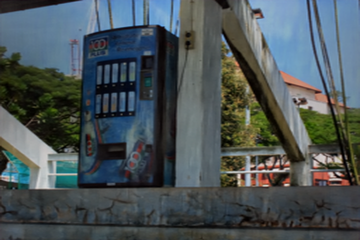}& \hspace{-0.4cm}
				\includegraphics[width = 0.15\textwidth ]{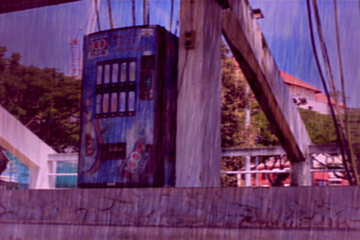}& \hspace{-0.4cm}
				\includegraphics[width = 0.15\textwidth ]{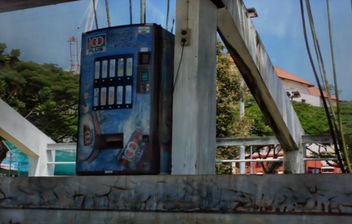}& \hspace{-0.4cm}
				\includegraphics[width = 0.15\textwidth ]{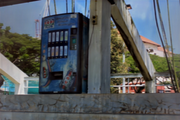}& \hspace{-0.4cm}
				\includegraphics[width = 0.15\textwidth ]{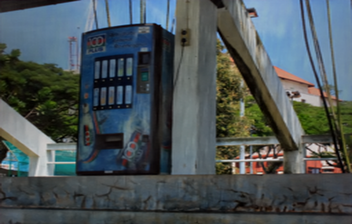}& \hspace{-0.4cm}
				\includegraphics[width = 0.15\textwidth ]{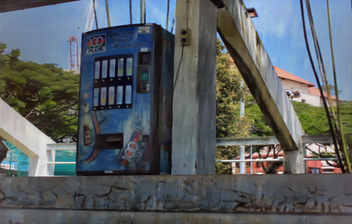}& \hspace{-0.4cm}
				\includegraphics[width = 0.15\textwidth ]{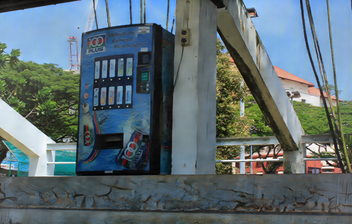}& \hspace{-0.4cm}
				\includegraphics[width = 0.15\textwidth ]{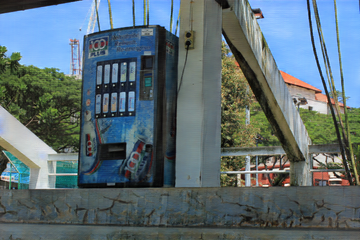}& \hspace{-0.4cm}
				\includegraphics[width = 0.15\textwidth ]{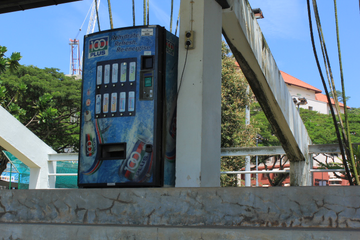}

				\\
				\includegraphics[width = 0.15\textwidth ]{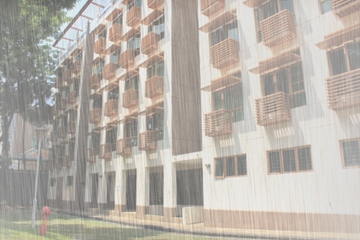}& \hspace{-0.4cm}
				\includegraphics[width = 0.15\textwidth ]{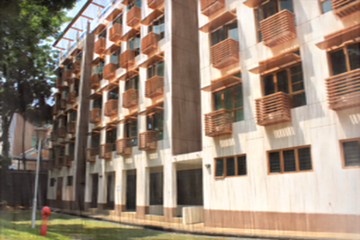}& \hspace{-0.4cm}
				\includegraphics[width = 0.15\textwidth ]{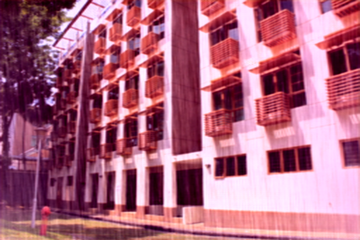}& \hspace{-0.4cm}
				\includegraphics[width = 0.15\textwidth ]{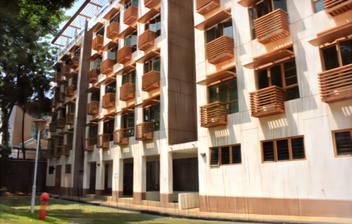}& \hspace{-0.4cm}
				\includegraphics[width = 0.15\textwidth ]{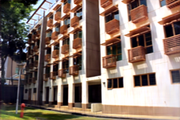}& \hspace{-0.4cm}
				\includegraphics[width = 0.15\textwidth ]{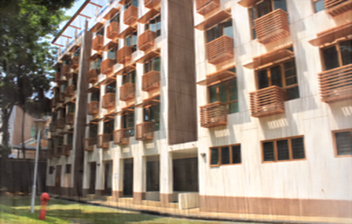}& \hspace{-0.4cm}
				\includegraphics[width = 0.15\textwidth ]{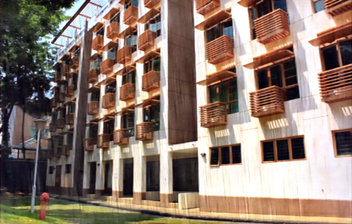}& \hspace{-0.4cm}
				\includegraphics[width = 0.15\textwidth ]{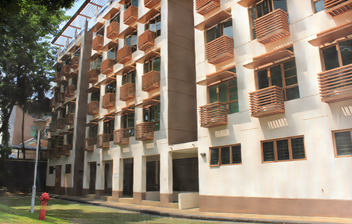}& \hspace{-0.4cm}
				\includegraphics[width = 0.15\textwidth ]{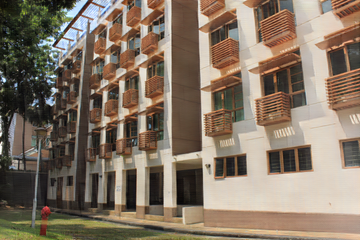}& \hspace{-0.4cm}
				\includegraphics[width = 0.15\textwidth ]{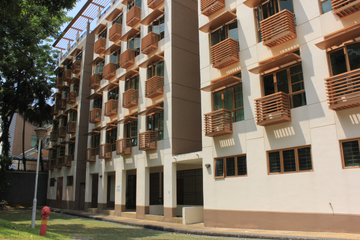}

				\\
				\includegraphics[width = 0.15\textwidth ]{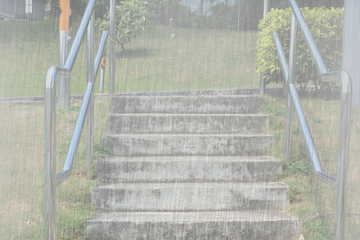}& \hspace{-0.4cm}
				\includegraphics[width = 0.15\textwidth ]{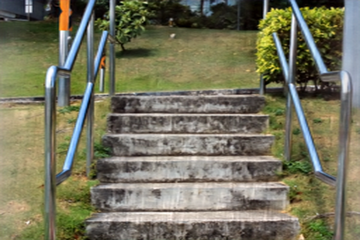}& \hspace{-0.4cm}
				\includegraphics[width = 0.15\textwidth ]{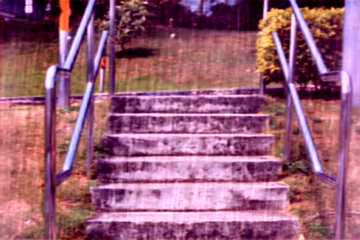}& \hspace{-0.4cm}
				\includegraphics[width = 0.15\textwidth ]{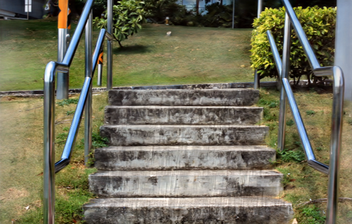}& \hspace{-0.4cm}
				\includegraphics[width = 0.15\textwidth ]{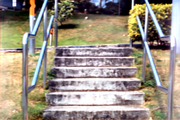}& \hspace{-0.4cm}
				\includegraphics[width = 0.15\textwidth ]{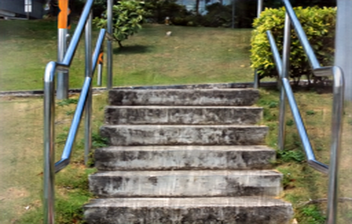}& \hspace{-0.4cm}
				\includegraphics[width = 0.15\textwidth ]{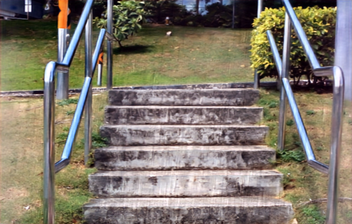}& \hspace{-0.4cm}
				\includegraphics[width = 0.15\textwidth ]{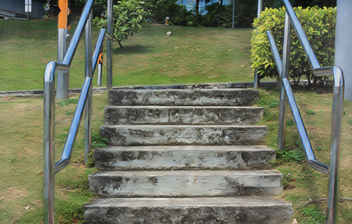}& \hspace{-0.4cm}
				\includegraphics[width = 0.15\textwidth ]{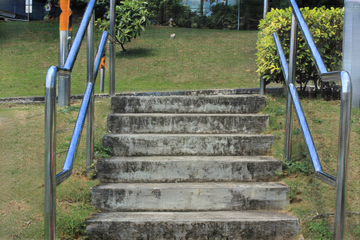}& \hspace{-0.4cm}
				\includegraphics[width = 0.15\textwidth ]{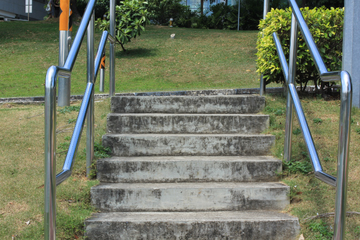}
				\\
				Input & \hspace{-0.4cm}
				DCSFN \cite{Wang2020}+\cite{Qin2020} & \hspace{-0.4cm}
				DCSFN \cite{Wang2020}+\cite{Shao2020} & \hspace{-0.4cm}
				MPRNet \cite{Zamir2021}+\cite{Qin2020} & \hspace{-0.4cm}
				MPRNet \cite{Zamir2021}+\cite{Shao2020} & \hspace{-0.4cm}
				NLEDN \cite{Li2018a}+\cite{Qin2020} & \hspace{-0.4cm}
				NLEDN \cite{Li2018a}+\cite{Shao2020} & \hspace{-0.4cm}
				HR \cite{Li2019a} & \hspace{-0.4cm} 
				Our & \hspace{-0.4cm}
				GT
				
			\end{tabular}
		}
	\end{center}
	\vspace{-0.3cm}
	\caption{Restoration results on our Rain-fog dataset and HeavyRain dataset \cite{Li2019a}.}
	\vspace{-0.35cm}
	\label{test1_derain_result}
\end{figure*}

\begin{table*}[t]
	\begin{center}	
		\caption{\\C{\scriptsize OMPARISON} R{\scriptsize ESULTS} {\scriptsize OF} A{\scriptsize VERAGE} PSNR, SSIM, AND LPIPS {\scriptsize ON} O{\scriptsize UR} R{\scriptsize AIN-FOG} D{\scriptsize ATASET} {\scriptsize ADN} H{\scriptsize EAVYRAIN} D{\scriptsize ATASET} \cite{Li2019a}}
		\label{test on test1}
		\begin{tabular}{r|c|ccc|cccccc}    
			\hline  
			\multicolumn{2}{c|}{Methods} & \multicolumn{3}{c|}{Rain-fog dataset} & \multicolumn{3}{c}{HeavyRain dataset \cite{Li2019a}}\\
			\hline
			\multicolumn{2}{c|}{Metric} & PSNR  & SSIM & LPIPS & PSNR  & SSIM & LPIPS \\    
			\hline
			\multirow{2}{*}{ DCSFN \cite{Wang2020} + FFA \cite{Qin2020} } &  DFDR & 12.253 & 0.251 & 0.601 & 14.064 & 0.651 & 0.369 
			\\  
			& DRDF & 11.404 & 0.281 & 0.600 & 15.359 & 0.660 & 0.214 
			\\
			\hline 
			\multirow{2}{*}{DCSFN \cite{Wang2020} + DA \cite{Shao2020} } & DFDR & 12.232 & 0.246 & 0.597 & 15.695  & 0.723  & 0.158 
			\\ 
			& DRDF & 11.257 & 0.254 & 0.602 & 15.305  & 0.655  & 0.391  
			\\ 
			\hline 
			\multirow{2}{*}{MPRNet \cite{Zamir2021} + FFA \cite{Qin2020}} & DFDR & 14.695 & 0.841 & 0.125 & 15.791 & 0.702 & 0.189  
			\\
			& DRDF & 15.629 & 0.721 & 0.159 & 14.738 & 0.662 & 0.189 
			\\ 
			\hline 
			\multirow{2}{*}{MPRNet \cite{Zamir2021} + DA \cite{Shao2020}} & DFDR & 19.117 & 0.765 & 0.131 & 16.149 & 0.765 & 0.131	
			\\
			& DRDF & 18.992 & 0.838 & 0.129 & 18.424 & 0.738 & 0.228 
			\\ 
			\hline 
			\multirow{2}{*}{NLEDN \cite{Li2018a} + FFA \cite{Qin2020}} 
			&  DFDR & 15.149 & 0.679 & \textbf{0.093} & 15.564 & 0.671 & 0.208
			\\&  DRDF & 16.076 & 0.763 & 0.133 & 15.359 & 0.660 & 0.214 \\
			\hline 
			\multirow{2}{*}{NLEDN \cite{Li2018a} + DA \cite{Shao2020}}
			&  DFDR & 18.284 & 0.809 & 0.128 & 21.264 & 0.809 & 0.128
			\\&  DRDF & 18.324 & 0.746 & 0.175 & 18.324 & 0.743 & 0.175 \\
			\hline 
			\multicolumn{2}{c|}{HR \cite{Li2019a}} & 21.126 & 0.806 & 0.130 & 22.182 & 0.849 & 0.083 \\
			\hline 
			\multicolumn{2}{c|}{Our} & \textbf{22.382} & \textbf{0.846} & 0.098 & \textbf{24.332} & \textbf{0.850} & \textbf{0.062}
			\\
			\hline     
		\end{tabular}  
		\vspace{-0.6cm}
	\end{center} 
\end{table*}

\subsection{Implementation Details}
Our ACGF is trained by the Pytorch 1.8.0 on a NVIDIA GeForce GTX 3060 GPU with 12GB memory. For training, a 256$ \times $256 image is randomly cropped from the original input size and normalizes the pixel values to [-1, 1]. Adam is employed as the optimization algorithm with a mini-batch size of 1. The model is trained for total 200 epochs. The learning rate starts from 0.0001 and decays with a policy of Pytorch after 100 epochs. Empirically initialize default values of $ \lambda_1 $, $ \lambda_2 $, $ \lambda_3 $ and $ \lambda_4 $ in Eq. (\ref{loss}) to 0.01, 10, 1, 1 respectively.

\subsection{Datasets, Comparison Methods and Evaluation Metric}
\subsubsection{Datasets}
We compare the proposed ACGF with state-of-the-art methods on six synthetic
rain datasets and one real-world rain dataset, include: (1) Since there are fewer datasets containing the fog, we synthesized a dataset based on the images from R100L \cite{Li2017}, named as Rain-fog dataset.
(2) HeavyRain dataset \cite{Li2019a} contains a total of 9000 rain-fog images. We select the last 1515 images as the test set, and the rest as the training set.
(3) Rain100L \cite{Li2017} has 200 pairs of light rain images for training and 100 pairs of images for testing.
(4) Rain800 \cite{Zhang2019} consists of 700 rain and clean image pairs for training and 100 pairs for testing.
(5) Rain12000 \cite{Zhang2018a} contains 12000 pairs of images for training. Moreover, 1200 synthetic image pairs are contained for test as well.
(6) Rain14000 \cite{Fu2017a} contains 14000 rain and clean image
pairs. They are synthesized from 1000 clean images with
14 kinds of different rain-streaks directions and scales. We select 12600 image pairs for training and the remaining 1400 pairs for testing.
(7) Real147 \cite{Wei2020} has 147 real-world rainy images without ground-truth.
We also compare our method with others on two synthetic fog datasets and two real fog datasets, which are:
(8) SOTS \cite{Li2018b} contains 500 indoor scenes and 500 outdoor scenes.
(9) HazeRD \cite{Zhang2017} contains 75 synthetic outdoor hazy images in different fog concentrations. 
(10) O-haze \cite{Ancuti2018} contains 45 outdoor real-world hazy images for testing.
(11) LIVE \cite{Choi2015} dataset contains 500 outdoor real-world hazy images.

\subsubsection{Comparison Methods and Evaluation Metrics}
In this work, we qualitatively compare our proposed method with nine top-performing deraining methods. These baseline deraining methods are DCSFN \cite{Wang2020}, MPRNet \cite{Zamir2021}, NLEDN \cite{Li2018a}, HR \cite{Li2019a},
DerainNet \cite{Fu2017}, SPANet \cite{Wang2019}, LPNet \cite{Fu2019} SEMI \cite{Wei2020} and CycleDerain \cite{Wei2021}). Moreover, we also evaluate our method with some defogging methods, such as FFA \cite{Qin2020}, DA \cite{Shao2020}, Grid \cite{Liu2019a}, DehazeNet \cite{Cai2016}, EPDN \cite{Qu2019} and GCA \cite{Chen2019}.
To quantitatively evaluate the restoration quality for each method, we use the commonly reference and reference-free evaluation metrics, i.e., as peak signal-to-noise ratio (PSNR), structural similarity (SSIM) \cite{Wang2004}, 
learned perceptual image patch similarity (LPIPS) \cite{Zhang2018},
naturalness image quality evaluator (NIQE) \cite{Mittal2012},
spatial-spectral entropy-based quality (SSEQ) \cite{Liu2014}
and blind/referenceless image spatial quality evaluator (BRISQE) \cite{Mittal2012a}. 

\subsection{Results on Synthetic Rain-fog Dataset}

Similar to \cite{Li2019a}, we combined several defogging (such as FFA \cite{Qin2020} and DA \cite{Shao2020}) and deraining (such as DCSFN \cite{Wang2020}, MPRNet \cite{Zamir2021} and NLEDN \cite{Li2018a}) methods in different order to evaluate the derained performance on Rain-fog and HeavyRain \cite{Li2019a} dateset. From defogging to deraining, we name as DFDR. From deraining to defogging, we denote as DRDF. 
In addition, we also compare our method with HR \cite{Li2019a}.
We use two full reference evaluation criteria: PSNR \cite{HuynhThu2008} and SSIM \cite{Wang2004} to evaluate the performance of all methods. 
To further evaluate deraining performance in perceptual level, we introduced another full reference evaluation index based on deep learning, LPIPS \cite{Zhang2018}. Different from the previous two evaluation indicators, the lower the value of LPIPS, the better the derain quality.
Quantitative results are shown in Table \ref{test on test1}.
It is clear that the proposed ACGF achieves significant improvements over these state-of-the-art methods. For the Rain-fog dataset, our method outperforms the over-leading method HR \cite{Li2019a} by 1.256 dB on PSNR, as well as 0.04 on SSIM. For the HeavyRain dataset \cite{Li2019a}, we can see that our method performs better than other methods on PSNR, SSIM and LPIPS metrics.
These results are consistent with the visual effects in Fig. \ref{test1_derain_result}. Although the competitive methods can remove the main rain streaks and fog from the rain-fog image, there is color distortion artifact. For example, as observed from the second image in Fig. \ref{test1_derain_result}, only our method restores the contrast and image content approximate to the ground truth. In contrast, other methods either have no effect on defogging performance, or appear obvious color distortion.
Moreover, compared with HR \cite{Li2019a}, our method removes rain streaks while recovering more detailed textures without artifacts. 
In contrast, as shown in the third and fifth columns of Fig. \ref{test1_derain_result}, other methods have different degrees of color cast phenomenon. 

\subsection{Results on Synthetic Rain Dataset}

\begin{figure*}[t]
	\begin{center}
		\resizebox{\textwidth}{!}
		{
			\begin{tabular}{@{}ccccccccc@{}}
				
				\includegraphics[width = 0.15\textwidth ]{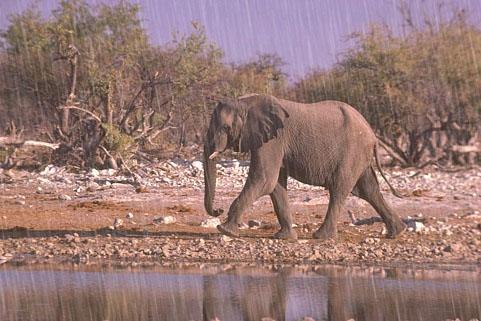}& \hspace{-0.4cm}
				\includegraphics[width = 0.15\textwidth ]{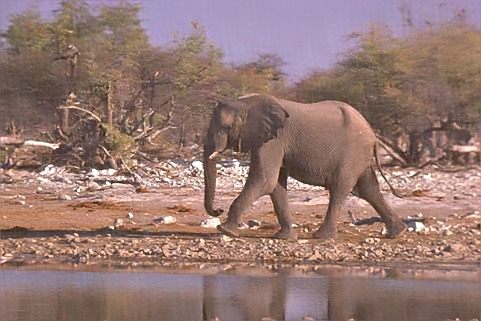}& \hspace{-0.4cm}
				\includegraphics[width = 0.15\textwidth ]{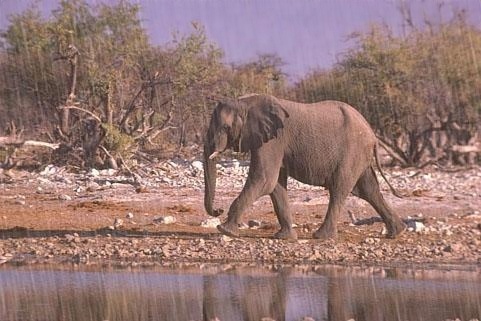}& \hspace{-0.4cm}
				\includegraphics[width = 0.15\textwidth ]{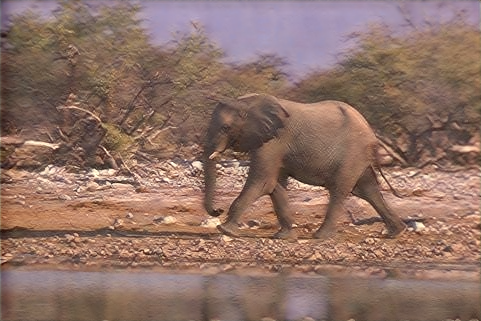}& \hspace{-0.4cm}
				\includegraphics[width = 0.15\textwidth ]{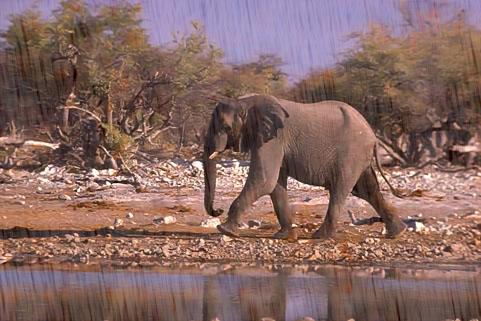}& \hspace{-0.4cm}
				\includegraphics[width = 0.15\textwidth ]{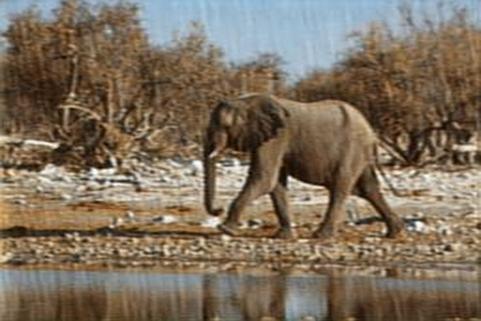}	& \hspace{-0.4cm}
				\includegraphics[width = 0.15\textwidth ]{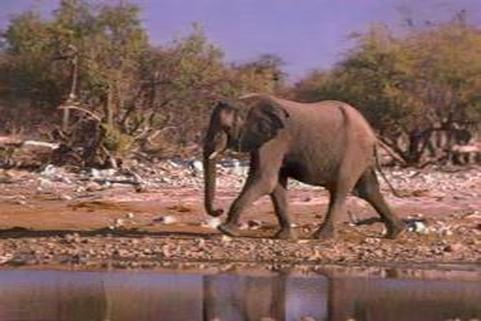}	& \hspace{-0.4cm}
				\includegraphics[width = 0.15\textwidth ]{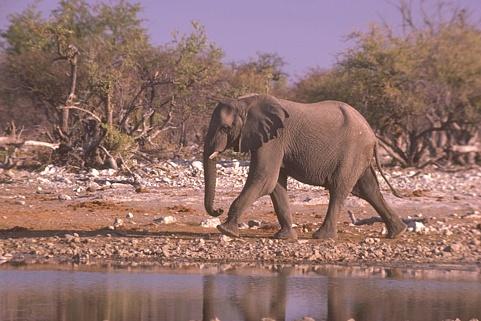}

				\\
				\includegraphics[width = 0.15\textwidth ]{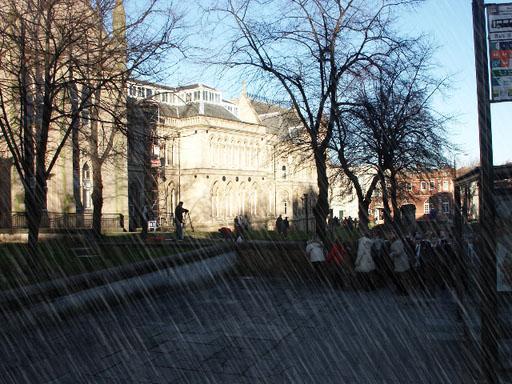}& \hspace{-0.4cm}
				\includegraphics[width = 0.15\textwidth ]{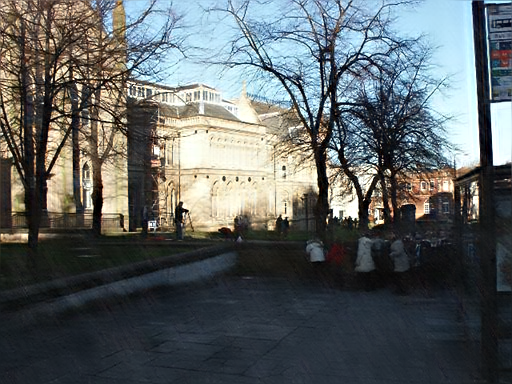}& \hspace{-0.4cm}
				\includegraphics[width = 0.15\textwidth ]{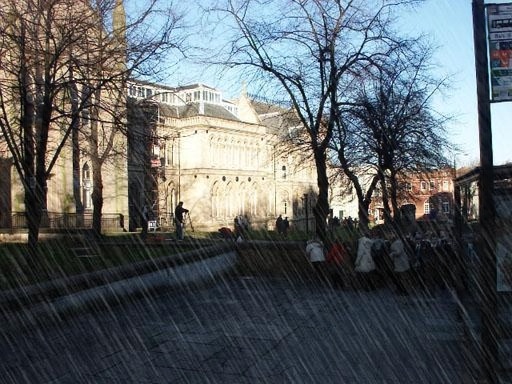}& \hspace{-0.4cm}
				\includegraphics[width = 0.15\textwidth ]{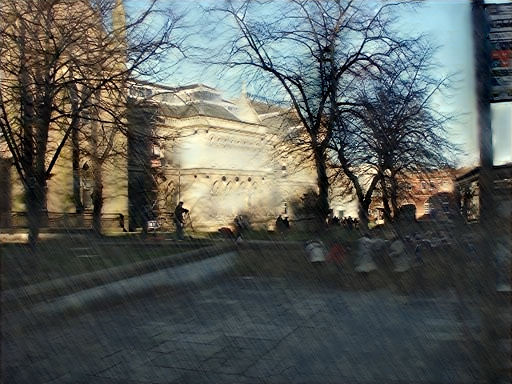}& \hspace{-0.4cm}
				\includegraphics[width = 0.15\textwidth ]{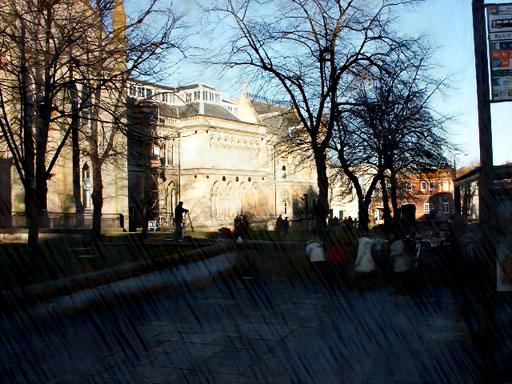}& \hspace{-0.4cm}
				\includegraphics[width = 0.15\textwidth ]{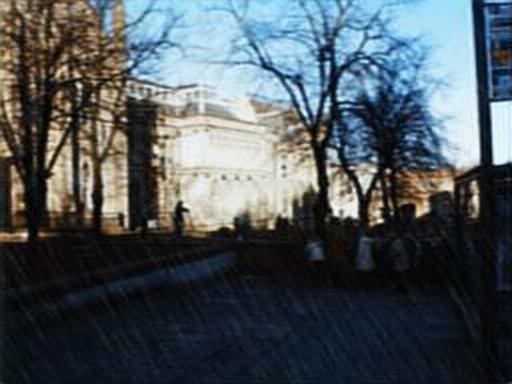}& \hspace{-0.4cm}
				\includegraphics[width = 0.15\textwidth ]{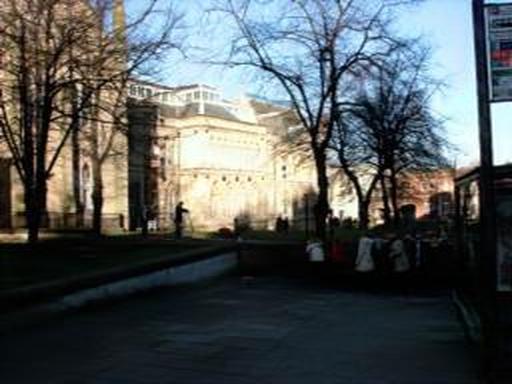}& \hspace{-0.4cm}
				\includegraphics[width = 0.15\textwidth ]{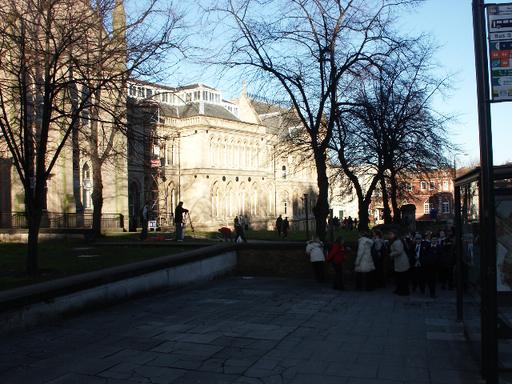}

				\\
				\includegraphics[width = 0.15\textwidth ]{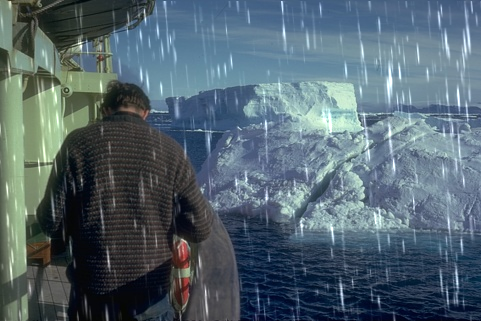}& \hspace{-0.4cm}
				\includegraphics[width = 0.15\textwidth ]{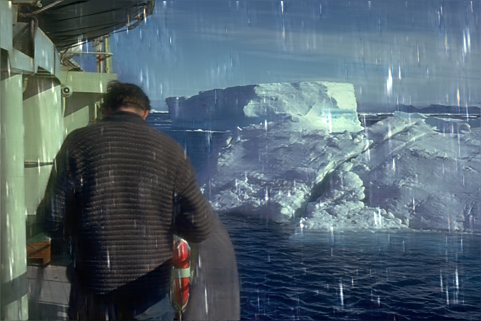}& \hspace{-0.4cm}
				\includegraphics[width = 0.15\textwidth ]{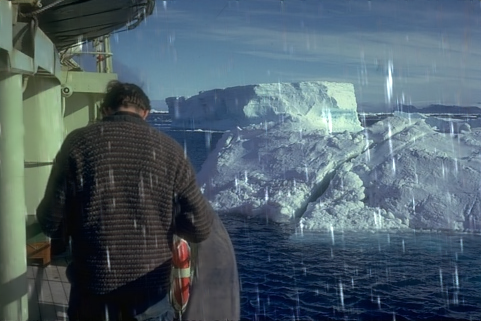}& \hspace{-0.4cm}
				\includegraphics[width = 0.15\textwidth ]{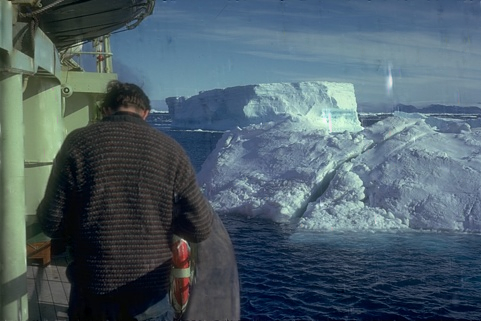}& \hspace{-0.4cm}
				\includegraphics[width = 0.15\textwidth ]{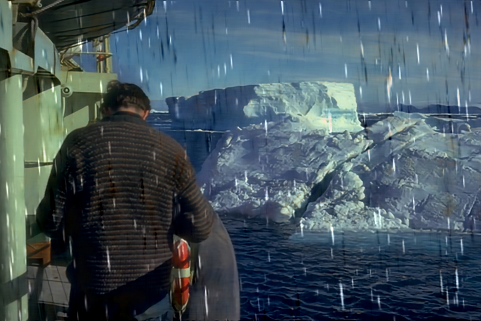}& \hspace{-0.4cm}
				\includegraphics[width = 0.15\textwidth ]{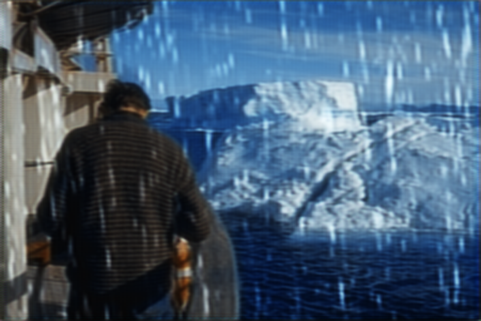}&  \hspace{-0.4cm}
				\includegraphics[width = 0.15\textwidth ]{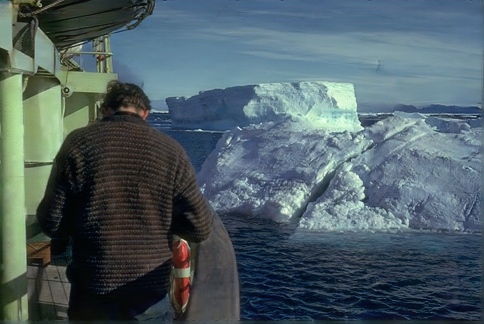}&  \hspace{-0.4cm}
				\includegraphics[width = 0.15\textwidth ]{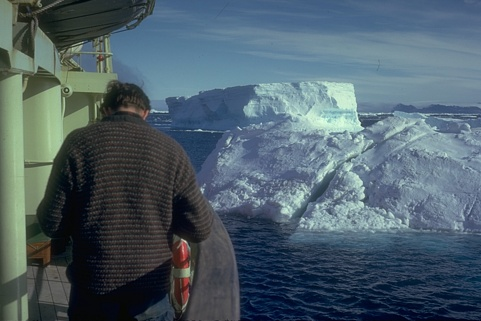}

				\\
				\includegraphics[width = 0.15\textwidth, height=0.15\textwidth]{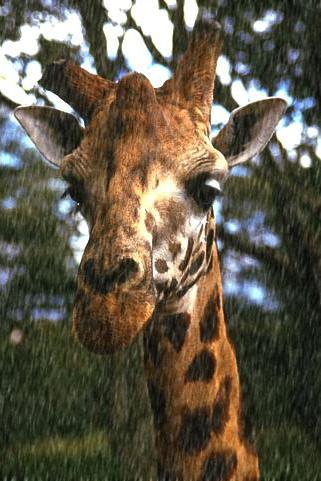}& \hspace{-0.4cm}
				\includegraphics[width = 0.15\textwidth, height=0.15\textwidth]{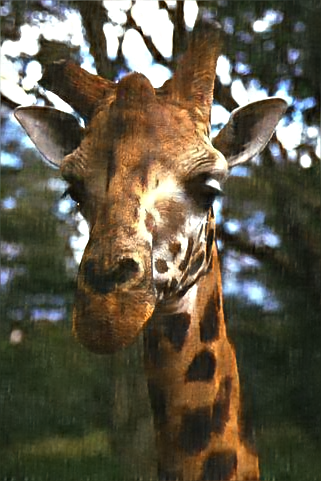}& \hspace{-0.4cm}
				\includegraphics[width = 0.15\textwidth, height=0.15\textwidth]{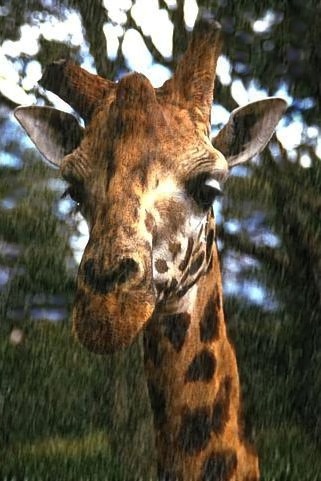}& \hspace{-0.4cm}
				\includegraphics[width = 0.15\textwidth, height=0.15\textwidth]{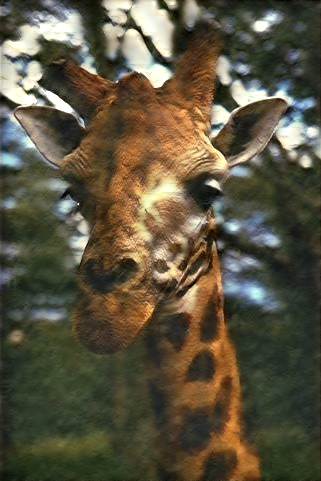}& \hspace{-0.4cm}
				\includegraphics[width = 0.15\textwidth, height=0.15\textwidth]{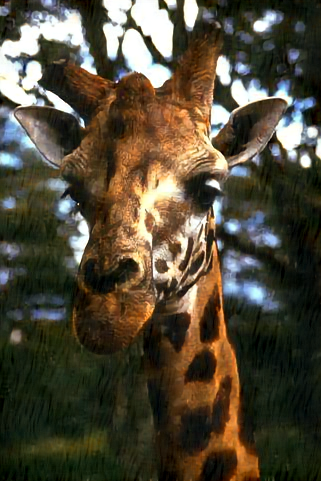}& \hspace{-0.4cm}
				\includegraphics[width = 0.15\textwidth, height=0.15\textwidth]{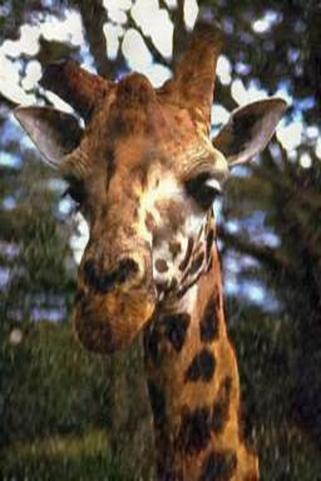}& \hspace{-0.4cm}			
				\includegraphics[width = 0.15\textwidth, height=0.15\textwidth]{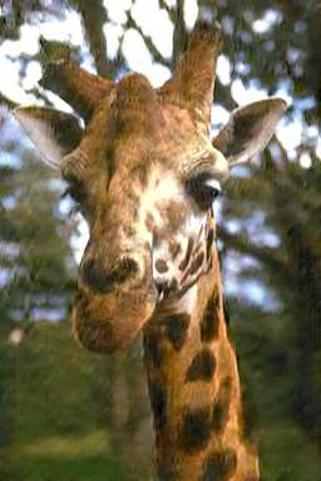}& \hspace{-0.4cm}
				\includegraphics[width = 0.15\textwidth, height=0.15\textwidth]{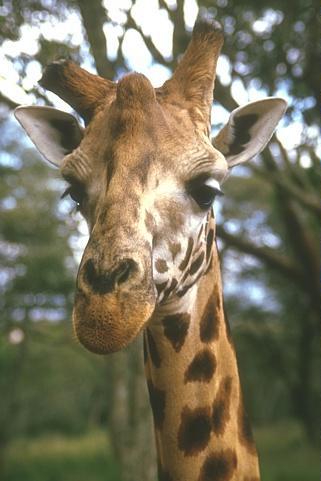}

				\\		
				\includegraphics[width = 0.15\textwidth ]{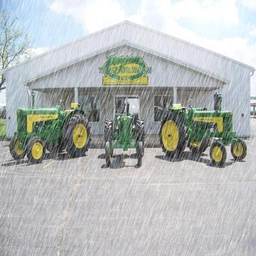}& \hspace{-0.4cm}
				\includegraphics[width = 0.15\textwidth ]{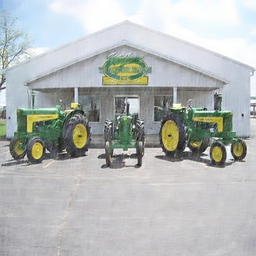}& \hspace{-0.4cm}
				\includegraphics[width = 0.15\textwidth ]{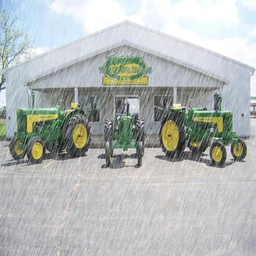}& \hspace{-0.4cm}
				\includegraphics[width = 0.15\textwidth ]{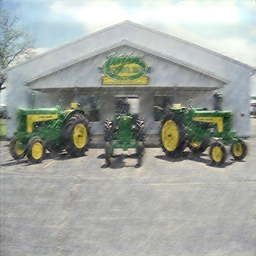}& \hspace{-0.4cm}
				\includegraphics[width = 0.15\textwidth ]{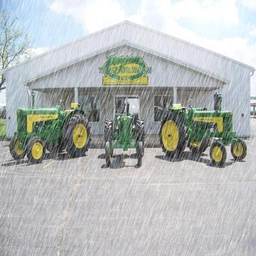}& \hspace{-0.4cm}
				\includegraphics[width = 0.15\textwidth ]{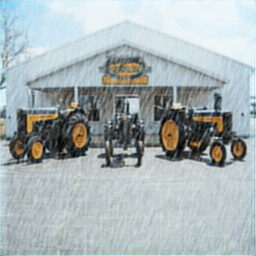}&  \hspace{-0.4cm}
				\includegraphics[width = 0.15\textwidth ]{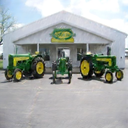}&  \hspace{-0.4cm}
				\includegraphics[width = 0.15\textwidth ]{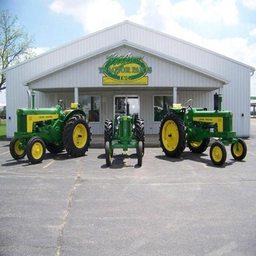}
				
				\\
				\includegraphics[width = 0.15\textwidth ]{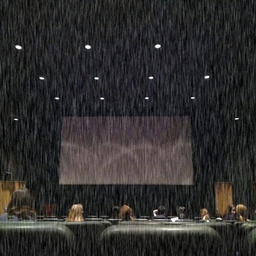}& \hspace{-0.4cm}
				\includegraphics[width = 0.15\textwidth ]{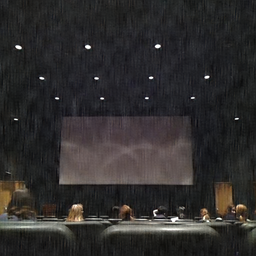}& \hspace{-0.4cm}
				\includegraphics[width = 0.15\textwidth ]{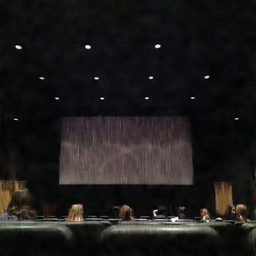}& \hspace{-0.4cm}
				\includegraphics[width = 0.15\textwidth ]{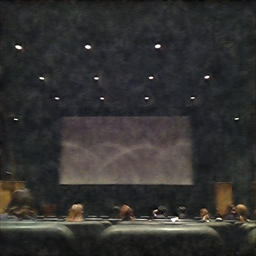}& \hspace{-0.4cm}
				\includegraphics[width = 0.15\textwidth ]{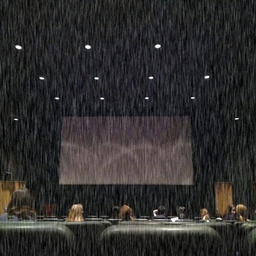}& \hspace{-0.4cm}
				\includegraphics[width = 0.15\textwidth ]{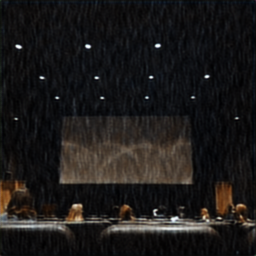}& \hspace{-0.4cm}
				\includegraphics[width = 0.15\textwidth ]{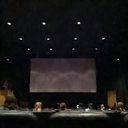}& \hspace{-0.4cm}
				\includegraphics[width = 0.15\textwidth ]{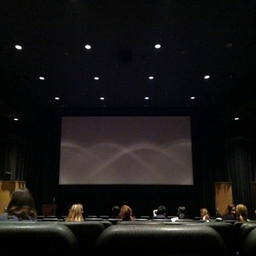}
				
				\\
				Input & \hspace{-0.4cm}
				DerainNet \cite{Fu2017} & \hspace{-0.4cm}
				SPANet \cite{Wang2019} & \hspace{-0.4cm}
				LPNet \cite{Fu2019}  & \hspace{-0.4cm}
				SEMI \cite{Wei2020} & \hspace{-0.4cm}
				CycleDerain \cite{Wei2021} & \hspace{-0.4cm}
				Our & \hspace{-0.4cm}
				GT	
			\end{tabular} %
		}
	\end{center}
	\vspace{-0.3cm}
	\caption{Restoration results on synthetic datasets, including Rain100L \cite{Li2017}, Rain800 \cite{Zhang2019}, Rain14000 \cite{Fu2017a}, and Rain12000 \cite{Zhang2018a}.}
	\label{syn_datasets derain}
	\vspace{-0.15cm}
\end{figure*}

\begin{table*}
	\begin{center}
		\caption{\\C{\scriptsize OMPARISON} R{\scriptsize ESULTS} {\scriptsize OF} A{\scriptsize VERAGE} PSNR, SSIM, {\scriptsize AND} LPIPS {\scriptsize ON} S{\scriptsize EVERAL} R{\scriptsize AIN} D{\scriptsize ATASETS},
			INCLUDING RAIN100L \cite{Li2017}, RAIN800 \cite{Zhang2019}, RAIN14000 \cite{Fu2017a}, AND RAIN12000 \cite{Zhang2018a}.
		}
		\label{syn derain}
		
		\begin{tabular}{lccccccccccc}    
			\toprule  
			\multirow{2}{*}{Methods}
			& Rain100L \cite{Li2017} & Rain800 \cite{Zhang2019} & Rain14000 \cite{Fu2017a} & Rain12000 \cite{Zhang2018a} \\  
			& PSNR/SSIM/LPIPS & PSNR/SSIM/LPIPS & PSNR/SSIM/LPIPS  & PSNR/SSIM/LPIPS  \\
			\midrule 
			
			DerainNet \cite{Fu2017} & 26.85/0.845/0.133 & 22.00/0.780/\textcolor{blue}{0.131} & 25.04/0.841/\textcolor{red}{0.062} & 21.53/\textcolor{blue}{0.798}/\textcolor{blue}{0.091} \\   
			SPANet \cite{Wang2019} & 28.46/0.921/0.128 & \textcolor{blue}{22.68}/\textcolor{blue}{0.787}/0.205 & 25.22/\textcolor{blue}{0.852}/0.127 & \textcolor{blue}{23.53}/0.770/0.159\\  
			LPNet \cite{Fu2019} & \textcolor{red}{33.28}/\textcolor{red}{0.943}/\textcolor{red}{0.032} & 20.92/0.756/0.188 & 22.03/0.765/0.174 & 22.22/0.781/0.127  \\
			SEMI \cite{Wei2020} &23.75/0.799/0.156 & 21.16/0.731/0.138 & \textcolor{blue}{26.34}/0.831/\textcolor{blue}{0.068} & 22.50/0.724/0.185\\
			CycleDerain \cite{Wei2021} & 20.15/0.647/0.379 & 21.75/0.721/0.210 & 21.07/0.673/0.333 & 21.30/0.694/0.308 \\
			Our &\textcolor{blue}{29.15}/\textcolor{blue}{0.927}/\textcolor{blue}{0.036} & \textcolor{red}{24.17}/\textcolor{red}{0.792}/\textcolor{red}{0.104}& \textcolor{red}{26.54}/\textcolor{red}{0.867}/0.107 & \textcolor{red}{27.743}/\textcolor{red}{0.857}/\textcolor{red}{0.070} \\
			\bottomrule 
		\end{tabular} 
		\vspace{-0.5cm}
	\end{center} 
\end{table*}

Our proposed method also has competitive deraining performance on synthetic rain datasets. We evaluate six top-performing deraining methods, including DerainNet \cite{Fu2017}, SPANet \cite{Wang2019}, LPNet \cite{Fu2019}, SEMI \cite{Wei2020} and CycleDerain \cite{Wei2021}, on four datasets, i.e., Rain100L \cite{Yang2017}, Rain800 \cite{Zhang2019}, Rain14000 \cite{Fu2017a} and Rain12000 \cite{Zhang2018a}. 
As can be seen from Fig. \ref{syn_datasets derain}, compared with other methods, our method can well remove the rain streaks from the image in different rain densities.
For instance, as observed from images in the third row of Fig. \ref{syn_datasets derain}, our proposed method restores clear image details and approximate contrast, which are closer to the ground truth. 
Although other methods can remove the rain streaks from the image, there are some rain streaks remain in the derained result (such as DerainNet \cite{Fu2017}, SPANet \cite{Wang2019}, SEMI \cite{Wei2020}, etc.) and have a serious color distortion (CycleDerain \cite{Wei2021}).
Moreover, we found that it is difficult for the supervised learning-based methods to learn the color changes between the original image and the ground truth, but our method can learn it well (As shown in the fourth row of Fig. \ref{syn_datasets derain}). The reason is that our proposed pyramid attention block is more sensitive to the image content information. 
These qualitative evaluation results are also reflected in the quantitative evaluation results of Table \ref{syn derain}. 
As can be seen from Table \ref{syn derain}, our method has achieved good scores for evaluation on Rain800 \cite{Zhang2019} and Rain12000 \cite{Zhang2018a}, which further demonstrate that our method can recover more natural and color vivid results from the image in different densities.

\begin{figure*}[t]
	\begin{center}
		\resizebox{\textwidth}{!}{
			\begin{tabular}{@{}cccccccccc@{}}
				
				\includegraphics[width = 0.15\textwidth ]{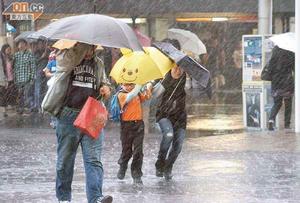}& \hspace{-0.4cm}
				\includegraphics[width = 0.15\textwidth ]{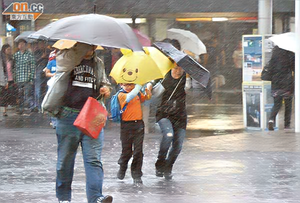}&\hspace{-0.4cm}
				\includegraphics[width = 0.15\textwidth ]{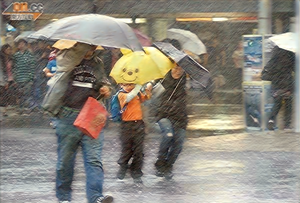}& \hspace{-0.4cm}
				\includegraphics[width = 0.15\textwidth ]{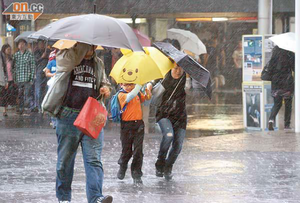}& \hspace{-0.4cm}
				\includegraphics[width = 0.15\textwidth ]{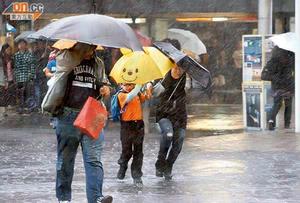}& \hspace{-0.4cm}
				\includegraphics[width = 0.15\textwidth ]{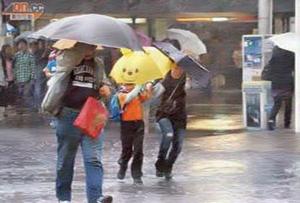}& \hspace{-0.4cm}
				\includegraphics[width = 0.15\textwidth ]{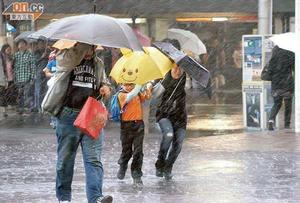}& \hspace{-0.4cm}

				\\
				\includegraphics[width = 0.15\textwidth ]{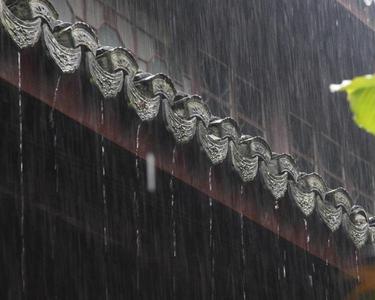}& \hspace{-0.4cm}
				\includegraphics[width = 0.15\textwidth ]{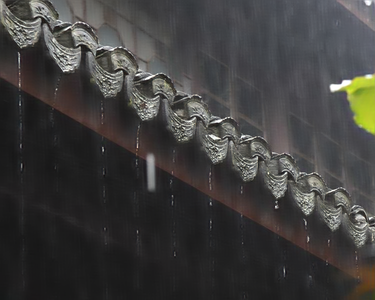}&\hspace{-0.4cm}
				\includegraphics[width = 0.15\textwidth ]{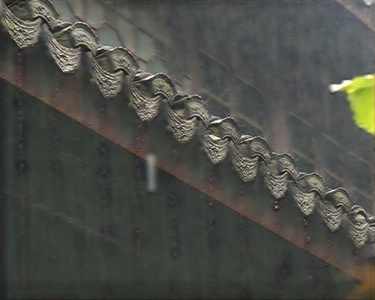}& \hspace{-0.4cm}
				\includegraphics[width = 0.15\textwidth ]{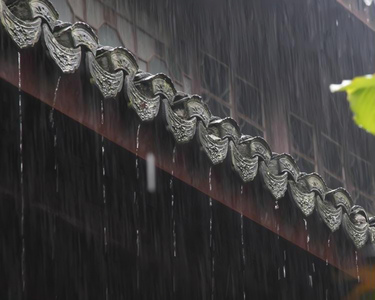}& \hspace{-0.4cm}
				\includegraphics[width = 0.15\textwidth ]{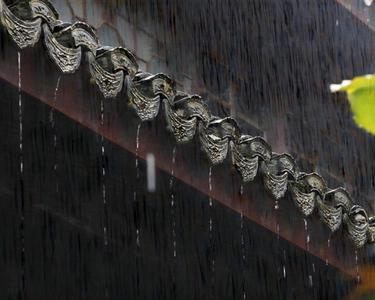}& \hspace{-0.4cm}
				\includegraphics[width = 0.15\textwidth ]{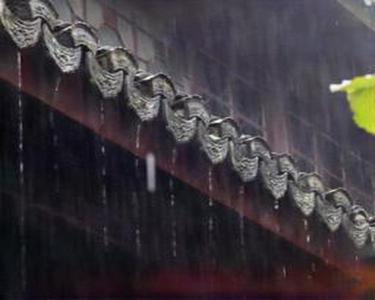}& \hspace{-0.4cm}
				\includegraphics[width = 0.15\textwidth ]{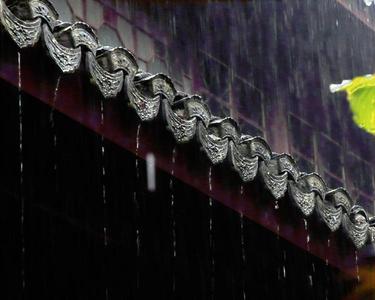}& \hspace{-0.4cm}
				
				\\		
				\includegraphics[width = 0.15\textwidth ]{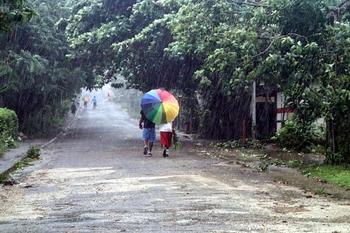}& \hspace{-0.4cm}
				\includegraphics[width = 0.15\textwidth ]{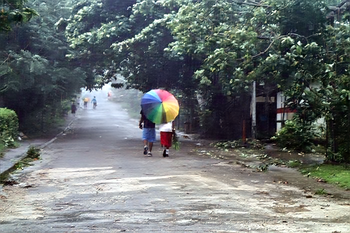}&\hspace{-0.4cm}
				\includegraphics[width = 0.15\textwidth ]{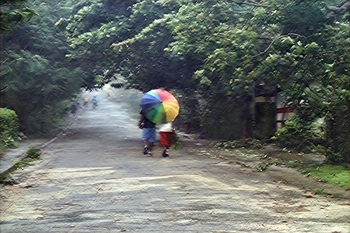}& \hspace{-0.4cm}
				\includegraphics[width = 0.15\textwidth ]{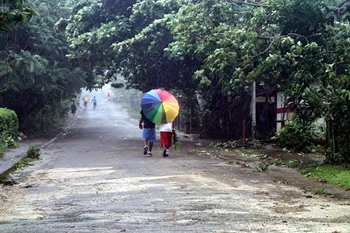}& \hspace{-0.4cm}
				\includegraphics[width = 0.15\textwidth ]{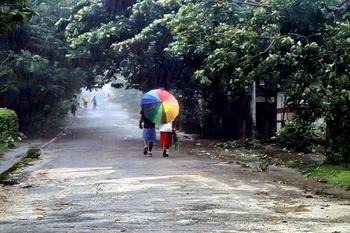}& \hspace{-0.4cm}
				\includegraphics[width = 0.15\textwidth ]{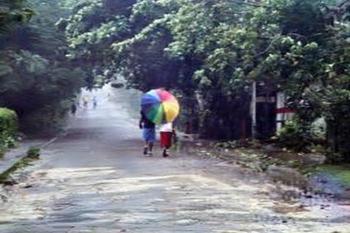}& \hspace{-0.4cm}
				\includegraphics[width = 0.15\textwidth ]{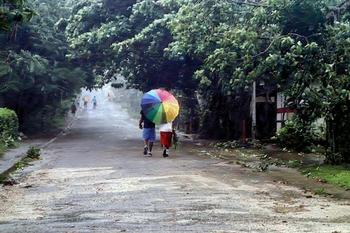}& \hspace{-0.4cm}
				
				\\
				\includegraphics[width = 0.15\textwidth ]{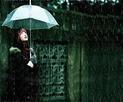}& \hspace{-0.4cm}
				\includegraphics[width = 0.15\textwidth ]{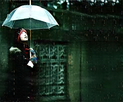}&\hspace{-0.4cm}
				\includegraphics[width = 0.15\textwidth ]{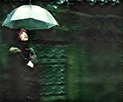}& \hspace{-0.4cm}
				\includegraphics[width = 0.15\textwidth ]{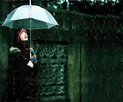}& \hspace{-0.4cm}
				\includegraphics[width = 0.15\textwidth ]{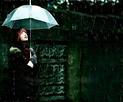}& \hspace{-0.4cm}
				\includegraphics[width = 0.15\textwidth ]{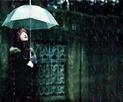}& \hspace{-0.4cm}
				\includegraphics[width = 0.15\textwidth ]{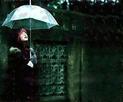}& \hspace{-0.4cm}
				
				\\
				Input & \hspace{-0.4cm}
				DerainNet \cite{Fu2017} & \hspace{-0.4cm}
				LPNet \cite{Fu2019}  & \hspace{-0.4cm}
				MPRNet \cite{Zamir2021}  & \hspace{-0.4cm}
				SEMI \cite{Wei2020} & \hspace{-0.4cm}
				CycleDerain \cite{Wei2021} & \hspace{-0.4cm}
				Our

			\end{tabular}
		}
	\end{center}
	\vspace{-0.3cm}
	\caption{Restoration results on real-world dataset \cite{Wei2020}.}
	\label{realworld_derain}
	\vspace{-0.35cm}
	
\end{figure*}

\subsection{Results on Real World Dataset}
\begin{figure*}[htbp]
	\begin{center}
		\resizebox{\textwidth}{!}{
			\begin{tabular}{@{}cccccccccc@{}}
				\\
				\includegraphics[width = 0.15\textwidth, height=0.09\textwidth]{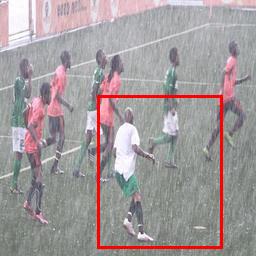}& \hspace{-0.4cm}
				\includegraphics[width = 0.15\textwidth, height=0.09\textwidth ]{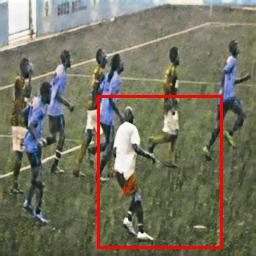}& \hspace{-0.4cm}
				\includegraphics[width = 0.15\textwidth, height=0.09\textwidth ]{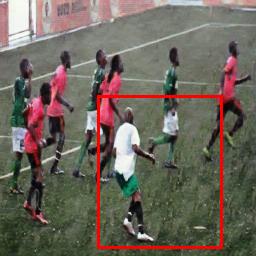}& \hspace{-0.4cm}	
				\includegraphics[width = 0.15\textwidth, height=0.09\textwidth ]{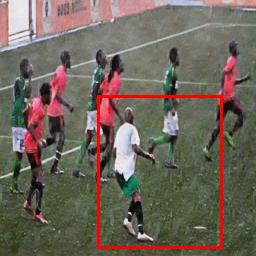}& \hspace{-0.4cm}
				\includegraphics[width = 0.15\textwidth, height=0.09\textwidth ]{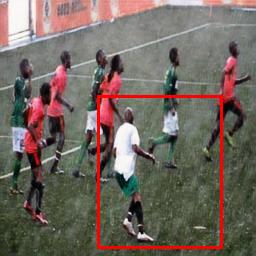}& \hspace{-0.4cm}
				\includegraphics[width = 0.15\textwidth, height=0.09\textwidth ]{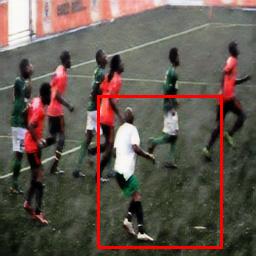}& \hspace{-0.4cm}
				\includegraphics[width = 0.15\textwidth, height=0.09\textwidth ]{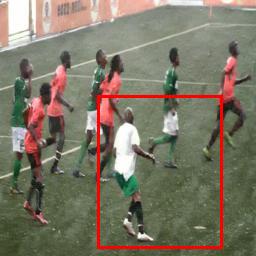}
				\\
				\includegraphics[width = 0.15\textwidth, height=0.09\textwidth]{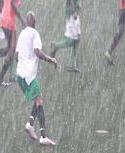}& \hspace{-0.4cm}
				\includegraphics[width = 0.15\textwidth, height=0.09\textwidth ]{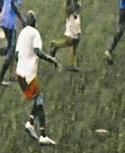}& \hspace{-0.4cm}
				\includegraphics[width = 0.15\textwidth, height=0.09\textwidth ]{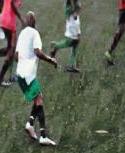}& \hspace{-0.4cm}	
				\includegraphics[width = 0.15\textwidth, height=0.09\textwidth ]{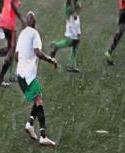}& \hspace{-0.4cm}
				\includegraphics[width = 0.15\textwidth, height=0.09\textwidth ]{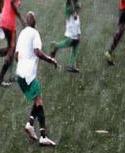}& \hspace{-0.4cm}
				\includegraphics[width = 0.15\textwidth, height=0.09\textwidth ]{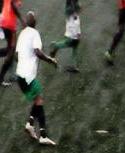}& \hspace{-0.4cm}
				\includegraphics[width = 0.15\textwidth, height=0.09\textwidth ]{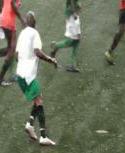}
				\\
				(a)  & \hspace{-0.4cm}
				(b)  & \hspace{-0.4cm}
				(c)  & \hspace{-0.4cm}
				(d)  & \hspace{-0.4cm}
				(e)  & \hspace{-0.4cm}
				(f)  & \hspace{-0.4cm}
				(g)  
				
			\end{tabular}
		}
	\end{center}
	\vspace{-0.3cm}
	\caption{
		Ablation study on real-world dataset. (a) Input rain image. (b) The rain removal result by using S1 structure. (c) The rain removal result by using our ACGF without the perception loss and the loss respectively are removed. (d) The rain removal result by using our ACGF without the diverse loss respectively is removed. (e) The rain removal result by using our ACGF without ARFE, but has the same number of network parameters. (f) The rain removal result of by using our ACGF without PA, but has the same number of network parameters. (g) Our ACGF.
	}
	\label{ablation}
	\vspace{-0.35cm}
\end{figure*}

For most rain removal methods, it is still a challenge to recover details from the rain image in real-world environment. 
In this work, we evaluate our method on real-world dataset Real147 \cite{Wei2020} to further verify the generalization ability of our model. 
Since the real-world rainy image does not have the ground truth, two reference-free indicators, NIQE \cite{Mittal2012} and SSEQ \cite{Liu2014} are adopted to quantitatively evaluate the restoration performance. The smaller scores of SSEQ and NIQE indicate better deraining performance. Fig. \ref{realworld_derain} shows the comparison results on several representative real-world scenarios with DerainNet \cite{Fu2017}, LPNet \cite{Fu2017}, MPRNet \cite{Zamir2021}, SEMI \cite{Wei2020}, CycleDerain \cite{Wei2021} and our method.
We can see that, compared with our method, other methods can not remove the atmospheric veiling effect (as shown in the derained results in the third row of Fig. \ref{realworld_derain}) due to they disregard the influence of the fog. 
In Table \ref{compare on real rain}, we show the quantitative comparison results on the real-world dataset \cite{Wei2020}. For both the NIQE \cite{Mittal2012} and SSEQ \cite{Liu2014}, our method has the smallest values. It indicates that our proposed method exhibits more competitive visual performance than other methods in terms of texture details and color information.

\begin{table}[t]
	\begin{center}
		\caption{\\C{\scriptsize OMPARISON} R{\scriptsize ESULTS} {\scriptsize OF} A{\scriptsize VERAGE} NIQE/SSEQ
			{\scriptsize ON} {\scriptsize THE} R{\scriptsize EAL} W{\scriptsize ORLD} D{\scriptsize ATASET} \cite{Wei2020}.}
		\label{compare on real rain}
		\resizebox{0.5\textwidth}{!}
		{
		\begin{tabular}{lccccccccc}    
			\toprule   
			Methods & DerainNet \cite{Fu2017} & LPNet \cite{Fu2019} & MPRNet \cite{Zamir2021} & SEMI \cite{Wei2020} & CycleDerain \cite{Wei2021} & our\\    
			\midrule 
			
			NIQE & 4.276 & 4.831 & 3.960 & 3.751 & 4.361 & \textbf{3.732}\\   
			SSEQ & 28.607  & 31.509 & 29.839 & 30.423 & 34.146 & \textbf{28.581}\\  
			\bottomrule   
		\end{tabular} } 
		\vspace{-0.5cm}
	\end{center} 
	
\end{table}

\subsection{Ablation Studies}
\subsubsection{Ablation Study on Network Structure}
\begin{table}[t]
	\begin{center}
		
		\caption{\\C{\scriptsize OMPARISON} R{\scriptsize ESULTS} {\scriptsize OF} A{\scriptsize VERAGE} NIQE/SSEQ
			{\scriptsize ON} {\scriptsize THE} R{\scriptsize EAL} W{\scriptsize ORLD} D{\scriptsize ATASET} {\scriptsize FOR} D{\scriptsize IFFERENT} C{\scriptsize ONFIGURATIONS}. \Checkmark M{\scriptsize EANS} {\scriptsize WITH}(W.) {\scriptsize THIS} M{\scriptsize ODULE}, \XSolidBrush M{\scriptsize EANS} W{\scriptsize ITHOUT}(W/O.) {\scriptsize THIS} M{\scriptsize ODULE}
		}
		\label{ablation study}
		
		\begin{tabular}{l|c|cccccccccc}    
			\hline 
			& S1 & \multicolumn{5}{c}{ACGF}  
			\\
			\hline
			ARFE    & \Checkmark &\Checkmark &\Checkmark & \XSolidBrush &\Checkmark &\Checkmark \\
			PA  & \Checkmark & \Checkmark & \Checkmark & \Checkmark & \XSolidBrush &\Checkmark  \\
			$ \mathcal{L}_{per} $ & \Checkmark & \XSolidBrush &\Checkmark & \Checkmark &\Checkmark &\Checkmark  \\
			$ \mathcal{L}_{div} $ & \Checkmark & \XSolidBrush & \XSolidBrush & \Checkmark & \Checkmark &\Checkmark\\
			\hline
			NIQE & 4.831 & 4.276 & 3.951 & 4.456 &4.557 &\textbf{3.732}  \\   
			SSEQ & 33.871  & 29.607 & 29.056 & 30.125& 36.253 &\textbf{28.581}  \\  
			\hline   
		\end{tabular} 
	\end{center} 
	\vspace{-0.7cm}
\end{table}

\begin{figure*}[t]
	\begin{center}
		\resizebox{\textwidth}{3cm}{
			\begin{tabular}{@{}cccccccccc@{}}
				\\
				\includegraphics[width = 0.15\textwidth ]{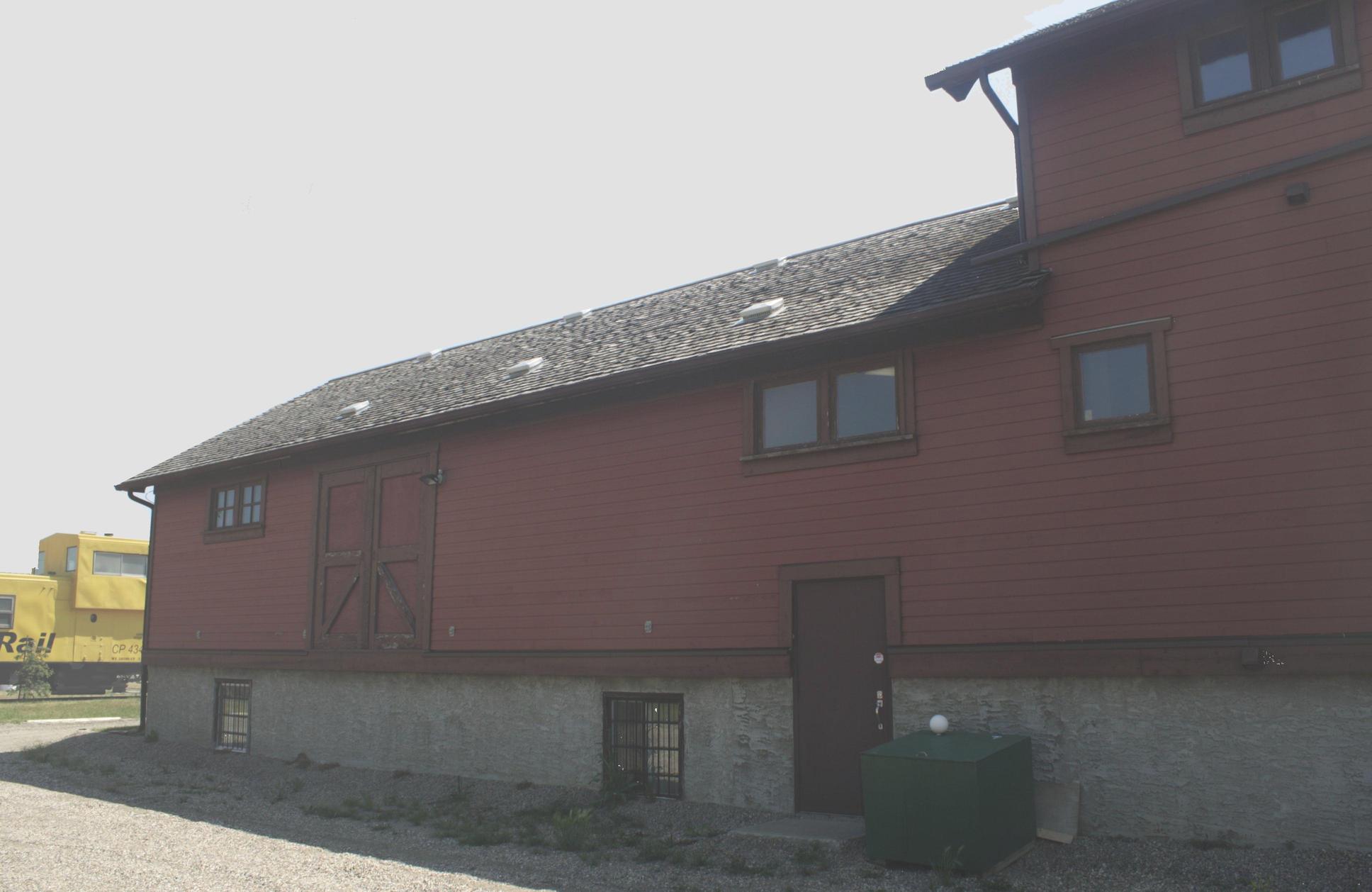}& \hspace{-0.4cm}
				\includegraphics[width = 0.15\textwidth ]{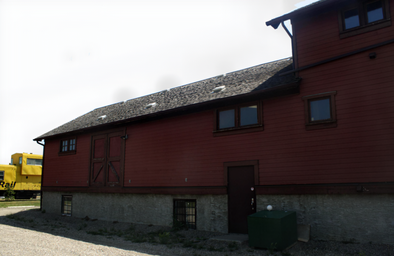}& \hspace{-0.4cm}
				\includegraphics[width = 0.15\textwidth ]{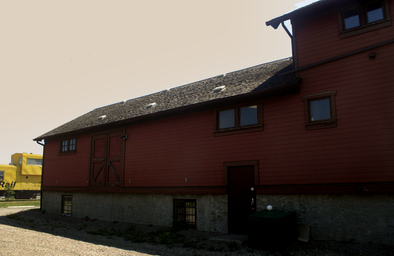}& \hspace{-0.4cm}
				\includegraphics[width = 0.15\textwidth ]{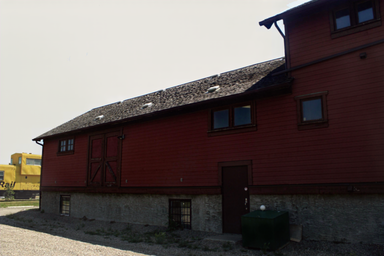}& \hspace{-0.4cm}
				\includegraphics[width = 0.15\textwidth ]{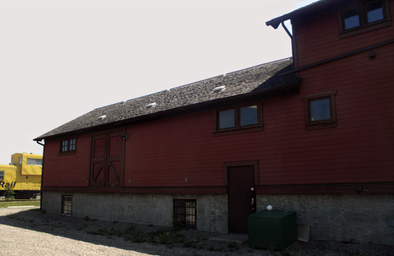}& \hspace{-0.4cm}
				\includegraphics[width = 0.15\textwidth ]{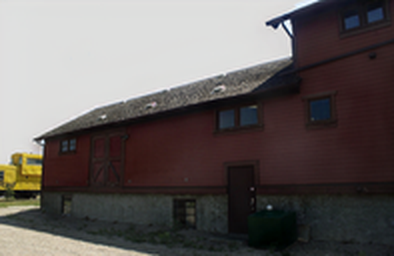}& \hspace{-0.4cm}
				\includegraphics[width = 0.15\textwidth ]{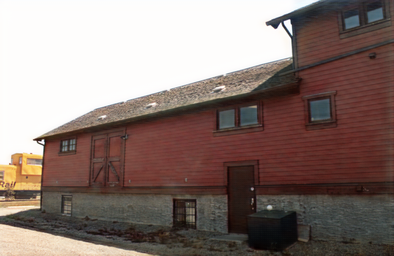}& \hspace{-0.4cm}
				\includegraphics[width = 0.15\textwidth ]{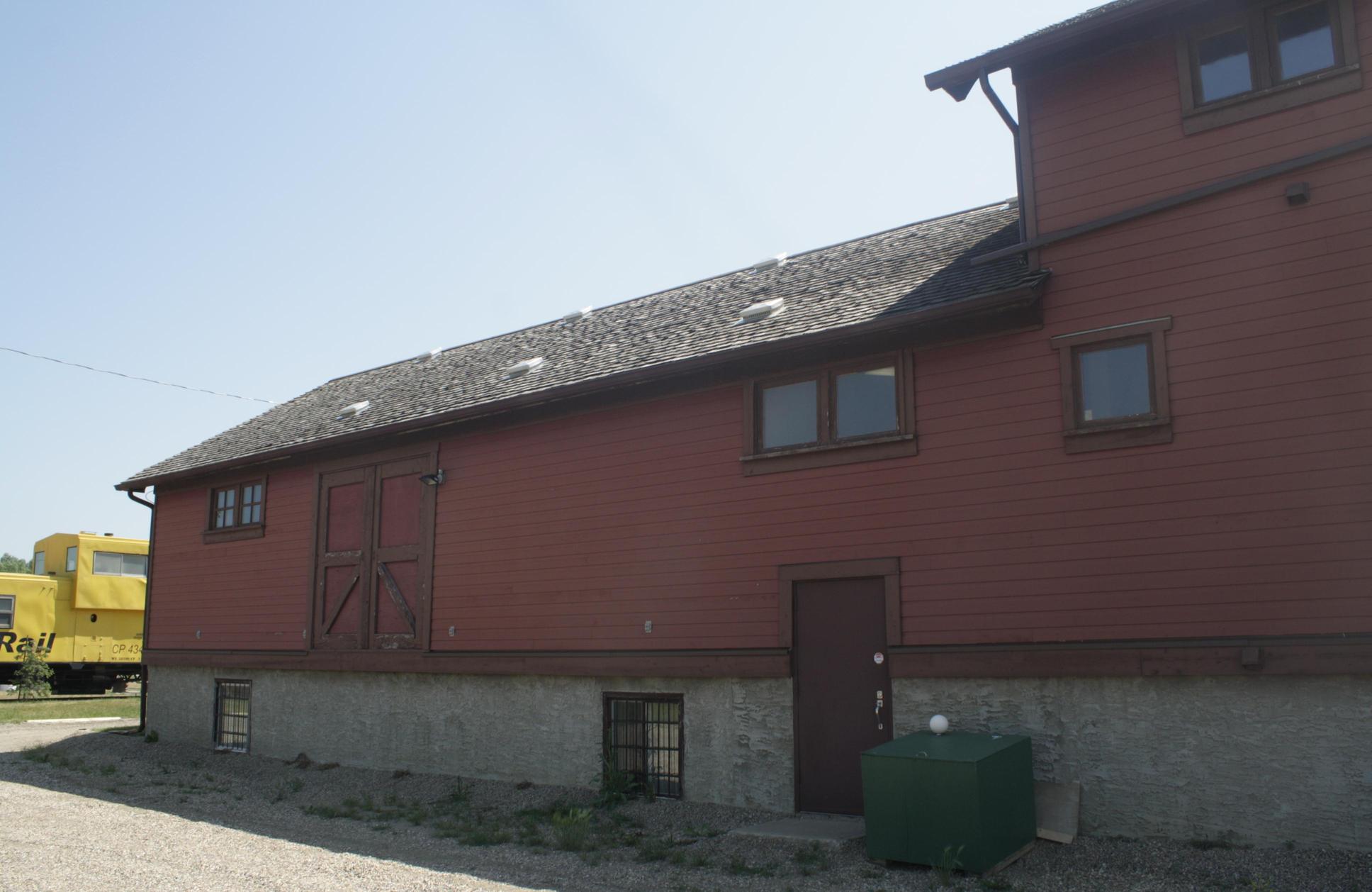}

				\\
				\includegraphics[width = 0.15\textwidth, height=0.09\textwidth ]{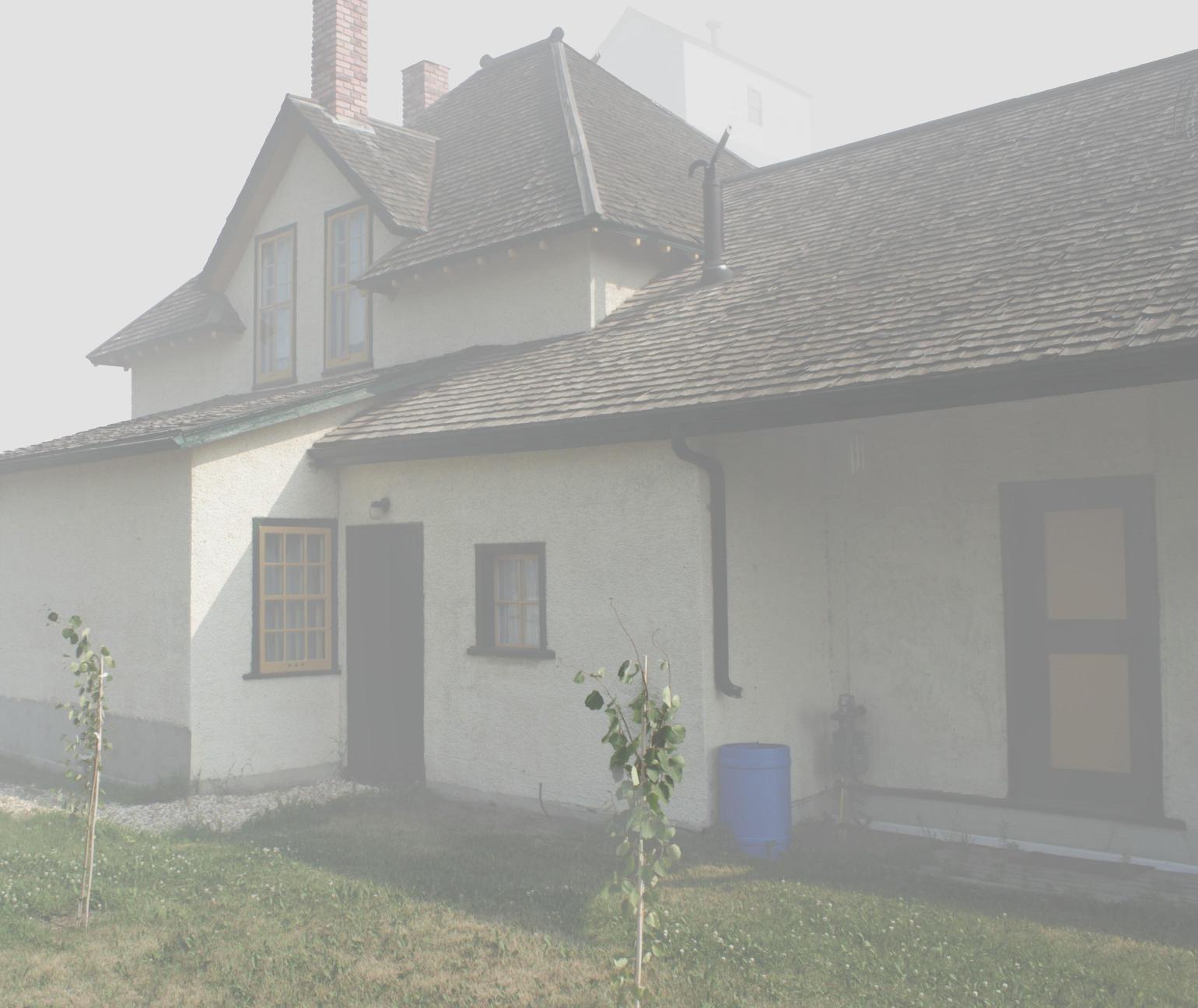}& \hspace{-0.4cm}
				\includegraphics[width = 0.15\textwidth, height=0.09\textwidth ]{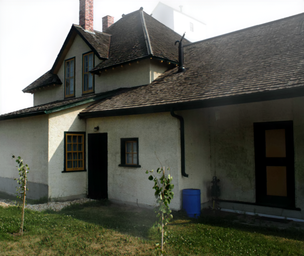}& \hspace{-0.4cm}
				\includegraphics[width = 0.15\textwidth, height=0.09\textwidth ]{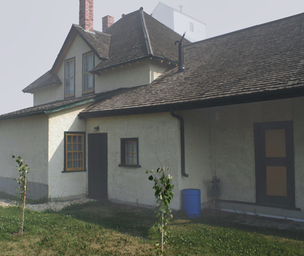}& \hspace{-0.4cm}
				\includegraphics[width = 0.15\textwidth, height=0.09\textwidth]{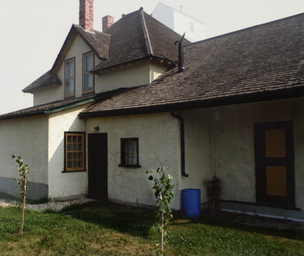}& \hspace{-0.4cm}
				\includegraphics[width = 0.15\textwidth, height=0.09\textwidth]{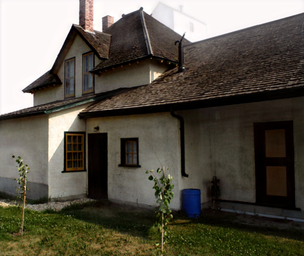}& \hspace{-0.4cm}
				\includegraphics[width = 0.15\textwidth, height=0.09\textwidth]{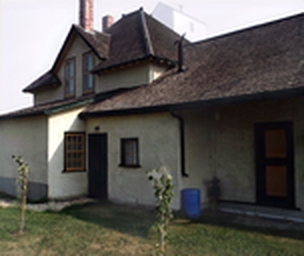}& \hspace{-0.4cm}
				\includegraphics[width = 0.15\textwidth, height=0.09\textwidth]{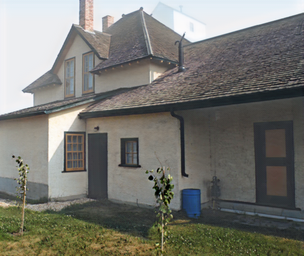}& \hspace{-0.4cm}
				\includegraphics[width = 0.15\textwidth, height=0.09\textwidth]{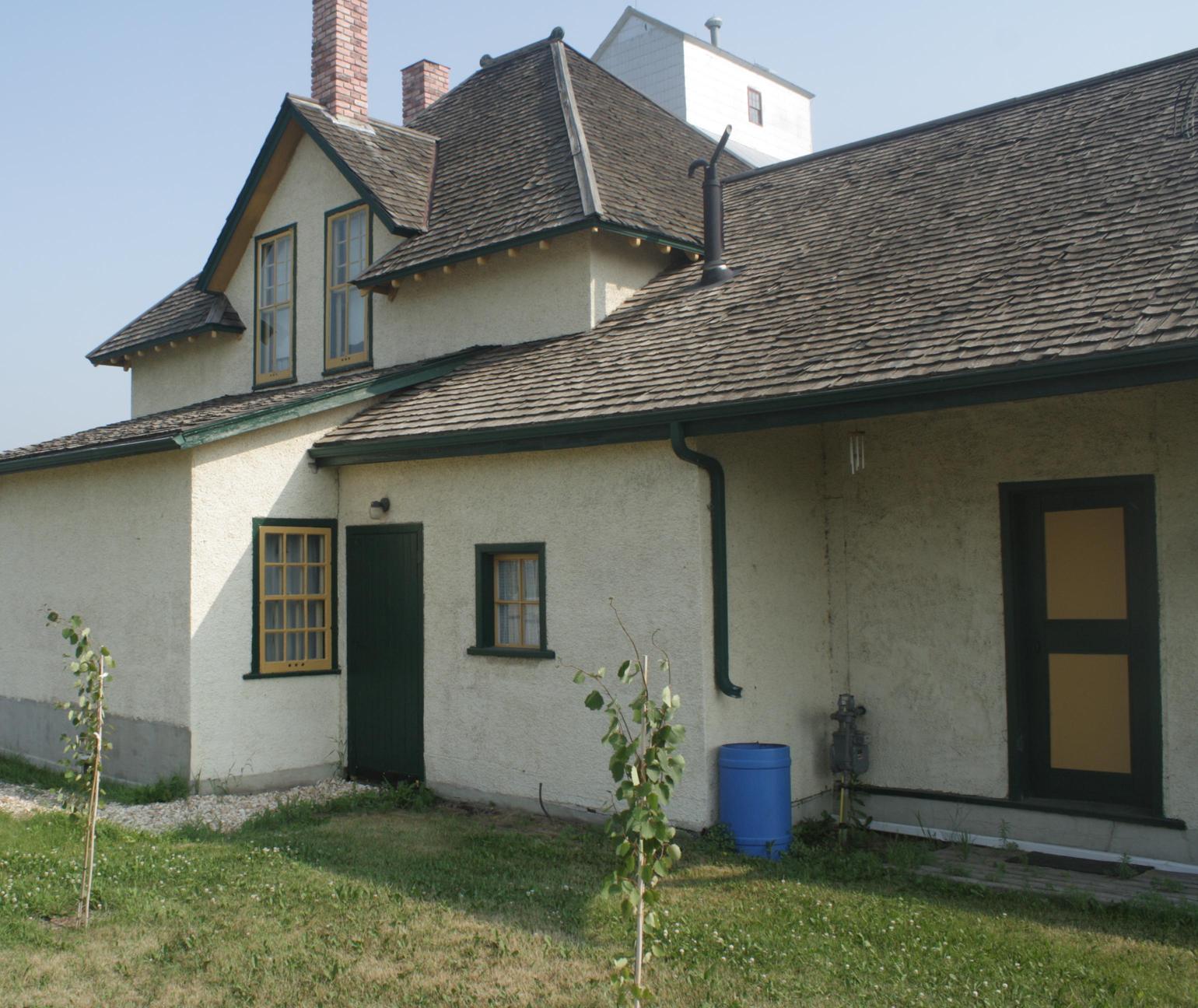}
				
				\\
				\includegraphics[width = 0.15\textwidth, height=0.09\textwidth]{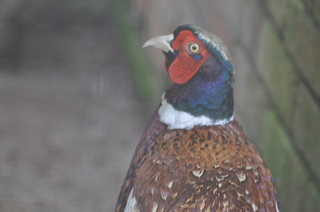}& \hspace{-0.4cm}
				\includegraphics[width = 0.15\textwidth, height=0.09\textwidth]{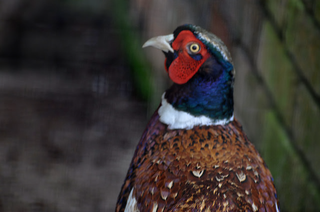}& \hspace{-0.4cm}
				\includegraphics[width = 0.15\textwidth, height=0.09\textwidth]{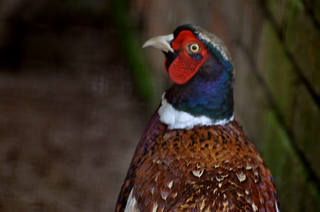}& \hspace{-0.4cm}
				\includegraphics[width = 0.15\textwidth, height=0.09\textwidth]{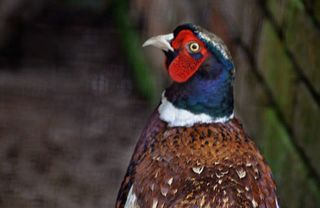}& \hspace{-0.4cm}
				\includegraphics[width = 0.15\textwidth, height=0.09\textwidth]{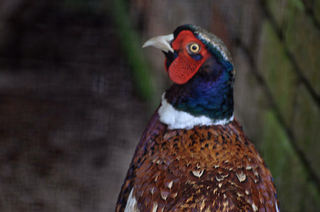}& \hspace{-0.4cm}
				\includegraphics[width = 0.15\textwidth, height=0.09\textwidth]{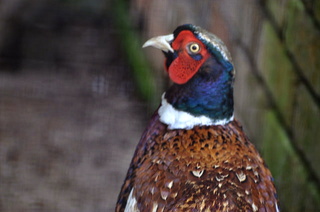}& \hspace{-0.4cm}
				\includegraphics[width = 0.15\textwidth, height=0.09\textwidth]{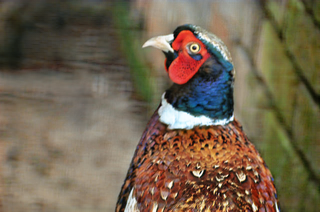}& \hspace{-0.4cm}
				
				\\
				\includegraphics[width = 0.15\textwidth, height=0.09\textwidth]{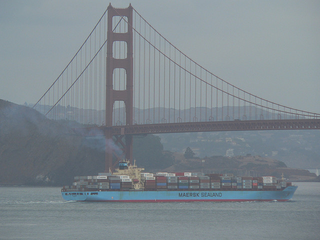}& \hspace{-0.4cm}
				\includegraphics[width = 0.15\textwidth, height=0.09\textwidth]{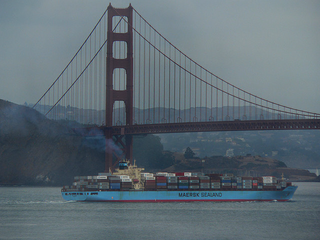}& \hspace{-0.4cm}
				\includegraphics[width = 0.15\textwidth, height=0.09\textwidth]{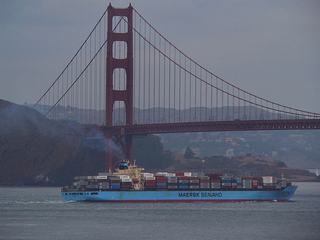}& \hspace{-0.4cm}
				\includegraphics[width = 0.15\textwidth, height=0.09\textwidth]{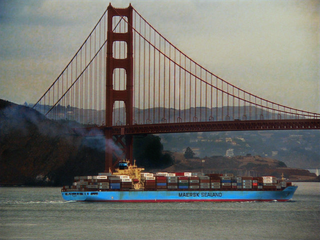}& \hspace{-0.4cm}
				\includegraphics[width = 0.15\textwidth, height=0.09\textwidth]{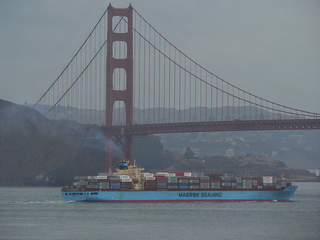}& \hspace{-0.4cm}
				\includegraphics[width = 0.15\textwidth, height=0.09\textwidth]{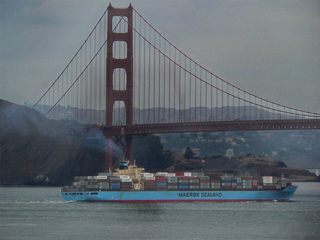}& \hspace{-0.4cm}
				\includegraphics[width = 0.15\textwidth, height=0.09\textwidth]{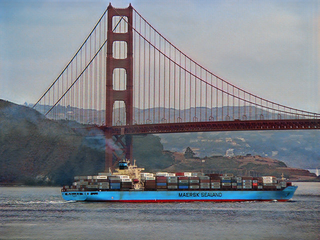}& \hspace{-0.4cm}

				\\
				Input & \hspace{-0.4cm}
				Grid \cite{Liu2019a} & \hspace{-0.4cm}
				DehazeNet \cite{Cai2016}  & \hspace{-0.4cm}
				EPDN \cite{Qu2019}  & \hspace{-0.4cm}
				FFA \cite{Qin2020} & \hspace{-0.4cm}
				GCA \cite{Chen2019} & \hspace{-0.4cm}
				Our & \hspace{-0.4cm}
				GT

			\end{tabular}
		}
	\end{center}
	\vspace{-0.3cm}
	\caption{Restoration results on synthetic fog datasets, including SOTS \cite{Li2018b}, HazeRD \cite{Zhang2017} and LIVE \cite{Choi2015}}
	\label{defog}
	\vspace{-0.5cm}
\end{figure*}

\begin{table*}[t]
	\begin{center}
		\caption{\\C{\scriptsize OMPARISON} R{\scriptsize ESULTS} {\scriptsize OF} A{\scriptsize VERAGE} NIQE {\scriptsize AND} BRISQUE {\scriptsize ON} S{\scriptsize EVERAL} F{\scriptsize OG} D{\scriptsize ATASETS},
			INCLUDING SOTS \cite{Li2018b}, HazeRD \cite{Zhang2017}, O-haze \cite{Ancuti2018}, AND LIVE \cite{Choi2015}.}
		\label{defog_result}
		\begin{tabular}{lcccccccccccc}    
			\toprule 
			\multirow{2}{*}{Methods} &
			SOTS \cite{Li2018b} & HazeRD \cite{Zhang2017} & O-haze \cite{Ancuti2018} & LIVE \cite{Choi2015} \\  
			&NIQE/BRISQUE  & NIQE/BRISQUE & NIQE/BRISQUE  & NIQE/BRISQUE  \\
			
			\midrule 
			Grid \cite{Liu2019a} & 3.421/11.151 & 4.096/15.415 & 2.830/15.937 & 4.096/12.701 \\   
			DehazeNet \cite{Cai2016} & 3.358/13.502 & 4.470/28.656 & \textcolor{blue}{2.617}/\textcolor{red}{11.630} & 4.470/16.818\\  
			EPDN \cite{Qu2019} & 3.673/\textcolor{blue}{10.958} & 4.134/19.107 & 3.673/18.132 & 4.134/\textcolor{blue}{9.779}  \\
			FFA \cite{Qin2020} &\textcolor{blue}{3.326}/11.314 & \textcolor{blue}{4.087}/24.696 & 3.326/15.973 & 4.087/12.670\\
			GCA \cite{Chen2019} & 3.568/15.814 & \textcolor{red}{3.715}/\textcolor{red}{14.555} & 3.568/21.712 & \textcolor{red}{3.715}/10.564 \\
			Our & \textcolor{red}{3.208}/\textcolor{red}{10.944} & 4.274/\textcolor{blue}{15.152} & \textcolor{red}{2.596}/\textcolor{blue}{15.540} & \textcolor{blue}{3.868}/\textcolor{red}{9.292} \\
			\bottomrule
		\end{tabular}  
	\end{center}
	\vspace{-0.2cm}
\end{table*}

To verify the effectiveness of our network structure, we first compare our model with the cascaded unsupervised learning framework (such as the structure of Fig. 1 (b)) that add our components (including ARFE, PA, Perceptual loss, and Diverse Loss), we denote it as S1.
Fig. \ref{ablation} (b) shows the derained result, and we can see that there is an obvious color cast phenomenon. Similarly, as can be seen from Table \ref{ablation study}, S1 has the highest values of NIQE and the second highest values of SSEQ, which means that this structure has the worst rain removal performance.

Second, we validate the contribution of each loss function by incrementally adding them to our base network model. Fig. \ref{ablation} (c) shows the derained result by using our model without Perceptual loss and Diverse loss. We can see that there are a lot of rain streaks that have not been removed and the vision blurred. Compared with the result by using our model without Diverse loss, although some rain streaks remain in the result, their texture details are clearer (as shown in Fig. \ref{ablation} (d)). It indicates that Perceptual loss can provide more image content information for the network layer to improve the quality of derained results.
Fig. \ref{ablation} (g) shows our ACGF's derained results, our method not only removes rain streaks and the fog, but also restores more details and generates higher fidelity images. Compared with the rain removal result without Diverse loss, the image generated by our ACGF is more natural without over enhancement.
This indicates that the Diverse loss helps the network to distinguish the rain image type and restore the image correctly according to the image type.
Table \ref{ablation study} also shows that the Diverse loss and Perceptual loss improve the quantitative results significantly. It can be reflected from the improvement of the objective evaluation index NIQE that the perceived quality of the generated image benefits from Diverse loss and Perceptual loss.

Finally, we verify the effectiveness of the ARFE and PA by replacing them with a network, but has the same number of parameters. As can be seen from Fig. \ref{ablation} (e), there are still some rain streaks that remain in the derained result by using our model without ARFE. Fig. \ref{ablation} (f) shows that the image details are lost in the derained result by using our model without PA. However, none of the above issues has occurred in our result (as shown in Fig. \ref{ablation} (g)). The reason is that our proposed ARFE can better learn the correlations between the global and local self-similarity of the rain-fog features, and PA can effectively separate the background and rain streaks information from rain images. Similarly, the results from Table VIII obviously demonstrate again the superiority of our proposed model. Compared with other results, our results have the best NIQE (3.732) and SSEQ (28.581).

\subsubsection{Comparison on Fog datasets}
To further demonstrate the effectiveness of our proposed model, we compare our method with several defogging methods (including Grid \cite{Liu2019a}, DehazeNet \cite{Cai2016}, EPDN \cite{Qu2019}, FFA \cite{Qin2020} and GCA \cite{Chen2019}) on different fog datasets (SOTS \cite{Li2018b}, HazeRD \cite{Zhang2017}, O-haze \cite{Ancuti2018}, and Live \cite{Choi2015}). As can be seen from Fig. \ref{defog}, although the competitive methods are able to remove haze in the input hazy images, the defogged results have the color distorted phenomenon (such as DehazeNet \cite{Cai2016} and EPDN \cite{Qu2019}) and low contrast (such as Grid \cite{Liu2019a} and FFA \cite{Qin2020}). In contrast, our method can recover the contrast and color information well for the defogged results. We also quantitatively use NIQE and BRISQUE to compare our method with other methods on several fog datasets. As shown in Table \ref{defog_result}, our method has a competitive defogging performance against other defogging methods.

\section{CONCLUSION}
In this article, we introduced a novel unsupervised attentive-adversarial learning framework (ACGF) for single image deraining, which simultaneously pays attention to the rain streaks and fog. ACGF consists of three major components: a Derain-fog network (DRFN), an Attention Rain-fog Feature Extraction network (ARFE), and a Rain-fog Feature Decoupling and Reorganization network (RFDR). Specifically, in ARFE, we exploit attentive and residual learning to learn the global similarity and local information complementarity of the rain relevant features (include rain streaks and fog), which can effectively capture the fusion rain-fog features from the rain image to improve the derained quality. Meanwhile, in RFDR, to improve the rain streaks removal ability of DRFN, we designed a pyramid attention block by using the high-level semantic information to compensate for the texture details loss caused by the low-level features. Moreover, to generate a more natural rainy image in RFDR to improve the generalization ability of our model, we proposed a diverse loss function to learn the difference among images in different domains. Extensive experiments of single image deraining on synthetic and real-world rain images show that our ACGF has a remarkable rain streaks removal performance against the state-of-the-art methods.

\ifCLASSOPTIONcaptionsoff
  \newpage
\fi



%

{\small
	\bibliographystyle{ieee}
	\bibliography{paper}
}

%








\end{document}